\def\year{2020}\relax
\documentclass[letterpaper]{article} % DO NOT CHANGE THIS
\usepackage{aaai20}  % DO NOT CHANGE THIS
\usepackage{times}  % DO NOT CHANGE THIS
\usepackage{helvet} % DO NOT CHANGE THIS
\usepackage{courier}  % DO NOT CHANGE THIS
\usepackage[hyphens]{url}  % DO NOT CHANGE THIS
\usepackage{graphicx} % DO NOT CHANGE THIS
\usepackage{graphicx}  % DO NOT CHANGE THIS
\usepackage{changes}
\usepackage{soul}
\usepackage{multirow}
\usepackage{amsmath}
\usepackage{booktabs}
\usepackage[super]{nth}
\usepackage{array}
\newcommand{\etal}{\textit{et al}. }
\newcommand{\ie}{\textit{i}.\textit{e}.}
\newcommand{\eg}{\textit{e}.\textit{g}.}

\newcolumntype{?}[1]{!{\vrule width #1}}

\title{Dynamic Instance Normalization for Arbitrary Style Transfer}
\author{Yongcheng Jing\textsuperscript{\rm 1}, Xiao Liu\textsuperscript{\rm 2}, Yukang Ding\textsuperscript{\rm 2}, Xinchao Wang\textsuperscript{\rm 3}, \Large \textbf{Errui Ding}\textsuperscript{\rm 2}, \\ \Large \textbf{Mingli Song\textsuperscript{\rm 1}\thanks{Corresponding author},} \Large \textbf{Shilei Wen\textsuperscript{\rm 2}}\\
\textsuperscript{\rm 1}Zhejiang University, \textsuperscript{\rm 2}Department of Computer Vision Technology (VIS), Baidu Inc., \textsuperscript{\rm 3}Stevens Institute of Technology\\
\{ycjing, brooksong\}@zju.edu.cn, \{liuxiao12, dingyukang, dingerrui, wenshilei\}@baidu.com, xinchao.wang@stevens.edu
}
 
\begin{document}

\maketitle

%===============================================================================|=======================================
\begin{abstract}

Prior normalization methods rely on affine transformations
to produce arbitrary image style transfers, of which the
parameters are computed in a pre-defined way.
{Such manually-defined nature eventually results in the
high-cost and shared encoders for both style and content encoding,
making style transfer systems cumbersome to be deployed in
resource-constrained environments like on the mobile-terminal side.}
In this paper, we propose a new and generalized normalization module,
termed as \emph{Dynamic Instance Normalization~(DIN)},
that allows for flexible and more efficient arbitrary style transfers.
Comprising an instance normalization and a dynamic convolution,
DIN encodes a style image into learnable convolution parameters,
upon which the content image is stylized.
Unlike conventional methods that use shared complex encoders to encode content and style,
the proposed DIN introduces a sophisticated style encoder,
yet comes with a compact and lightweight content encoder for fast inference.
Experimental results demonstrate that the proposed approach
yields very encouraging results on
challenging style patterns and, to our best knowledge,
for the first time enables an
arbitrary style transfer using
MobileNet-based lightweight
architecture,
%leading to a speedup factor of more than twenty
%as compared to existing approaches.
%\jyc{leading to a computational cost of one twentieth as compared to existing approaches.}
%\jyc{leading to a more than twenty times reduction in computational complexity as compared to existing approaches.}
{leading to a reduction factor of more than twenty in computational cost as compared to existing approaches.}
Furthermore, the proposed DIN provides
flexible support for state-of-the-art convolutional operations,
and thus triggers novel functionalities, such as
uniform-stroke placement for non-natural images
and automatic spatial-stroke control.
%\xwc{Shall we say something about the result improvement?}

\end{abstract}
%\animategraphics[height=2.5in,autoplay,loop,controls]{5}{figs/animate_}{0}{16}
%===============================================================================|=======================================
\section{Introduction}\label{section:Introduction}
%===============================================================================|=======================================
%
Image stylization has been a long-standing research topic.
It has been  studied in the domain of computer graphics, or more specifically, the area of \emph{Non-Photorealistic Rendering (NPR)} \cite{gooch2001non,rosin2012image}.
In the field of computer vision, image stylization is
studied as a generalized problem of texture synthesis \cite{efros1999texture}.
Built upon the recent progress in visual texture modelling \cite{gatys2015texture} and image reconstruction \cite{mahendran2015understanding}, Gatys \etal \cite{gatys2016image} propose to exploit \emph{Convolutional Neural Networks (CNNs)} to render a content image in different styles, pioneering a new field called \emph{Neural Style Transfer (NST)} \cite{jing2019neural}.

%------------------------------------------------------------------------------
%
\begin{figure}[!t]
\setlength\tabcolsep{1 pt}
{\renewcommand{\arraystretch}{0.6}
%\begin{tabular}{>{\centering}n{\p} >{\centering}n{\p} >{\centering\arraybackslash}n{\p}}
\begin{tabular}{>{\centering}m{2.05cm} >{\centering}m{2.05cm} >{\centering}m{2.05cm} >{\centering\arraybackslash}m{2.05cm}}
%\begin{tabular}{cccc}
\centering

%\includegraphics[width=0.115\textwidth]{figs/Quality/style4/content_img.jpg} & \includegraphics[width=0.115\textwidth]{figs/Quality/style4/nips.jpg} &\includegraphics[width=0.115\textwidth]{figs/Quality/style4/sheng.jpg}& \includegraphics[width=0.115\textwidth]{figs/Quality/style4/vgg.jpg} \\ \vspace{0.1cm}
%%\smallskip
%\scriptsize{Content}&  \scriptsize{Li \etal}&\scriptsize{Sheng \etal}& \scriptsize{Ours (VGG)} \\
%
% \includegraphics[width=0.115\textwidth]{figs/Quality/style4/style_img.jpg} & \includegraphics[width=0.115\textwidth]{figs/Quality/style4/swap.jpg} & \includegraphics[width=0.115\textwidth]{figs/Quality/style4/xun.jpg} &\includegraphics[width=0.115\textwidth]{figs/Quality/style4/unet.png}\\ \vspace{0.1cm}
%%\smallskip
%\scriptsize{Style} &\scriptsize{Chen \etal} &\scriptsize{Huang \etal} & \scriptsize{Ours (MobileNet)}

\includegraphics[width=0.1155\textwidth]{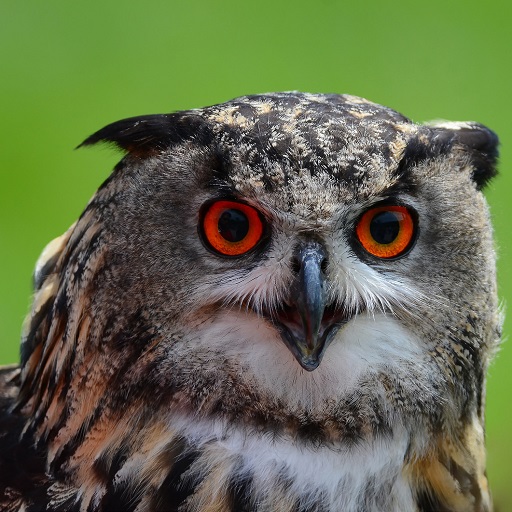}& \includegraphics[width=0.1155\textwidth]{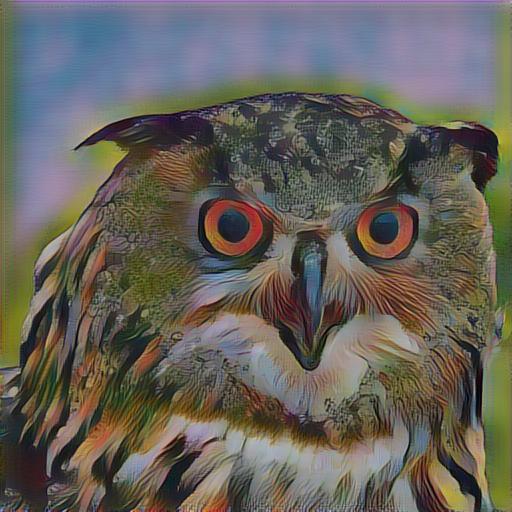} & \includegraphics[width=0.1155\textwidth]{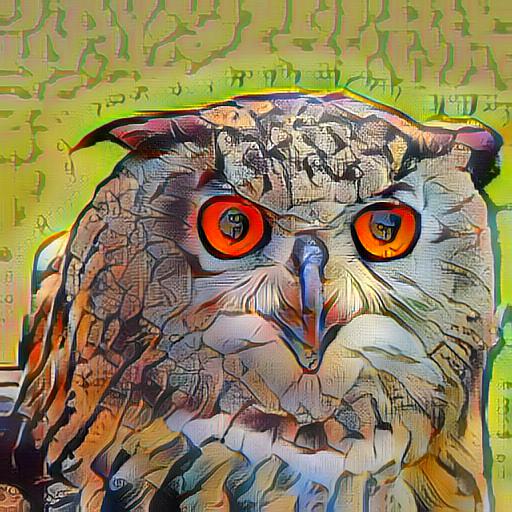}  & \includegraphics[width=0.1155\textwidth]{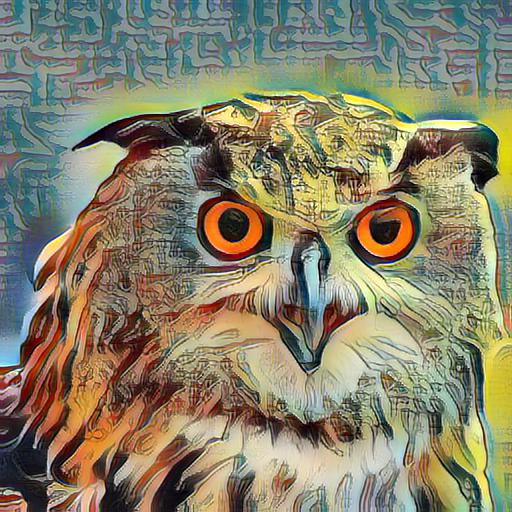} \\ %\vspace{0.1cm}
%\smallskip
\scriptsize{(a) Content}\vspace{0.1cm}&\scriptsize{(b) Chen \etal}\vspace{0.1cm} &\scriptsize{(c) Huang \etal}\vspace{0.1cm}& \scriptsize{(d) Ours~(VGG)}\vspace{0.1cm} \\

 \includegraphics[width=0.1155\textwidth]{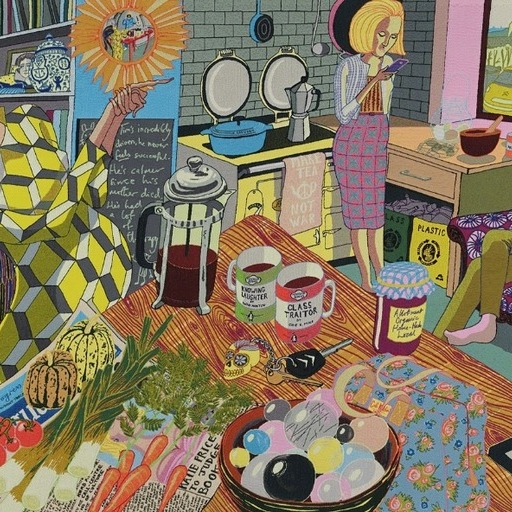} & \includegraphics[width=0.1155\textwidth]{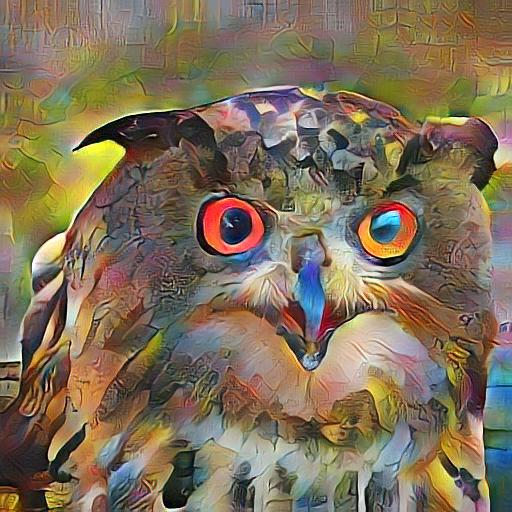} &\includegraphics[width=0.1155\textwidth]{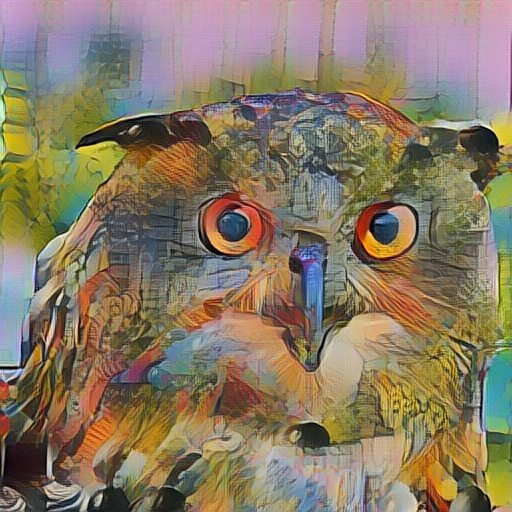}&\includegraphics[width=0.1155\textwidth]{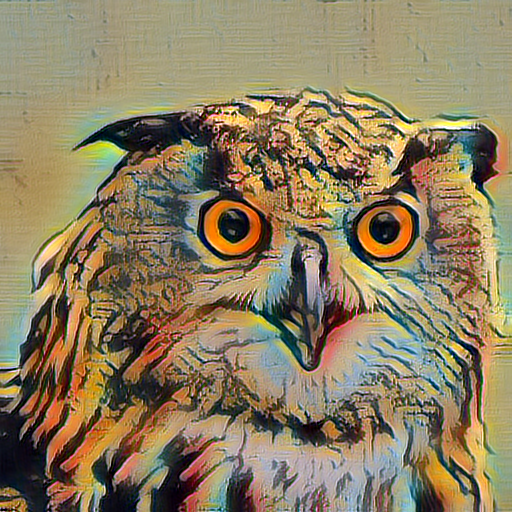}\\ %\vspace{0.1cm}
%\smallskip
\scriptsize{(e) Style} &  \scriptsize{(f) Li \etal}&\scriptsize{(g) Sheng \etal} & \vspace{-0.21cm} \multirow{1}{2.2cm}{\scriptsize{(h) Ours~(MobileNet)}}\\

%\scriptsize{(a)}&\scriptsize{(b)} & \scriptsize{(c)} & \scriptsize{(d)}
\end{tabular}
}
\vspace{-0.22cm}
%\vspace{-0.01cm}
\caption{Existing ASPM methods either barely
transfer style to the target~(\emph{Chen et al., Huang et al.}),
%\xwc{Why they ``fail''? be more specific}
or produce distorted style patterns (\emph{Li et al., Sheng et al.}) while relying on high-cost encoders. By contrast, the proposed DIN achieves superior performance using the same architecture (\emph{Ours~(VGG)}),
and for the first time endows a much smaller lightweight
network to transfer arbitrary styles (\emph{Ours~(MobileNet)}).}
\label{fig:showcase} %% label for entire figure
\end{figure}
%
%------------------------------------------------------------------------------

%Since the emergence of NST, it has been attracting increasing attention from both academia and industry \cite{jing2019neural}.
%
The inspiring work of Gatys~\etal is, however, built upon an iterative image optimization in the pixel space, which turns out to be
computationally expensive due to the online optimization.
%which is computationally expensive due to the burden of online optimization.
%
%Therefore, model-optimization-based NST algorithms are proposed to address this efficiency issue by optimizing feed-forward models in an offline training manner.
%To speed up the stylization process,
To address this efficiency issue, model-optimization-based NST algorithms are proposed, which optimize feed-forward models in an offline training manner.
%and then exploit their trained models to stylize an input image with one single forward pass during inference.
%
The earliest model-optimization-based NST algorithms, namely
\emph{Per-Style-Per-Model~(PSPM)},
train separate style-specific models for each particular style, and are therefore
burdensome to be adopted for real-world applications.
\cite{Johnson2016perceptual,ulyanov2016texture,li2016precomputed}.
To address this issue, \emph{Multiple-Style-Per-Model~(MSPM)} algorithms are proposed by incorporating multiple styles into one single model \cite{zhang2017multi,chen2017stylebank,li2017diverse,dumoulin2016learned}.
Unfortunately, MSPM also suffers from the inflexible
binding between specific styles and a single model.

More recently, \emph{Arbitrary-Style-Per-Model~(ASPM)} algorithms are proposed to solve the aforementioned dilemma by exploiting one single model to transfer arbitrary new styles \cite{chen2016fast,huang2017arbitrary,sheng2018neural,li2017universal}.
Despite the great progress, ASPM algorithms still
have limitations in handling complex style patterns and producing fine strokes.
Moreover,  they rely on high-cost and shared encoders
for both style and content encoding,
thus making the network cumbersome to be deployed on the mobile-terminal side.

In this paper, we propose a novel module, termed as
\emph{Dynamic Instance Normalization (DIN)},
that allows for flexible and
more efficient arbitrary style transfers.
Unlike existing approaches that require {a} pre-defined {way to compute}
parameters for their affine transformations for {arbitrary style transfer},
%\jyc{align the mean and variance of style features, the most basic statistics, to transfer arbitrary style,}
%{align the  basic statistics (\ie, mean and variance) of style features to transfer arbitrary style,}
%\jyc{[Comments: not sure whether it is confusing here.]}
%\jyc{align the most basic style feature statistics (\ie, mean and variance),}
%\jyc{align the feature mean and variance, the most basic statistics, between content and style,}
%\xwc{XXX say the function of the affine transformation},
the proposed DIN introduces a generalized
dynamic convolutional transformation,
of which the parameters are learned adaptively {for arbitrary stylization}.
%\jyc{for a more accurate alignment of the real complex statistics of style features}.
In this way,
%\jyc{with no need to align the feature mean and variance,}
%\jyc{with no need to align feature mean and variance,}
DIN makes it possible
to exploit a sophisticated style encoder to express
complex and rich style patterns,
and meanwhile preserves a
compact and lightweight content encoder for fast inference.
With the proposed DIN layer,
we are able to conduct arbitrary style transfer,
to our best knowledge, for the first time
on a compact MobileNet-based architecture,
resulting in plausible results with
%\jyc{a speedup factor of more than twenty times}
much lighter computational costs
as compared to
the state-of-the-art approaches.
Furthermore, DIN supports  various convolutional operations,
and therefore
enables novel transfer functionalities including
automatic spatial-stroke control and uniform-stroke
placement for non-natural images.

The proposed approach delivers gratifying visual results,
especially for finer strokes and sharper details.
An example that compares the visualizations of our
approach and those of other ASPM methods is shown in Fig.~\ref{fig:showcase},
where the goal is to transfer the artistic
style of a painting into a photo of an owl.
In the stylized results of Chen \etal and
Huang \etal (Fig.~\ref{fig:showcase}(b) and (c)),
the target style (Fig.~\ref{fig:showcase}(e))
is not well reflected, since few style patterns are transferred.
The results of Li \etal and Sheng \textit{et al}.~(Fig.~\ref{fig:showcase}(f) and (g)),
on the other hand, are prone to
distorted patterns and lack sharp details and fine strokes.
By contrast, the proposed method is able to parse
the challenging style patterns, like the wall brick
patterns in Fig.~\ref{fig:showcase}(e),
using the same architecture (Fig.~\ref{fig:showcase}(d)),
and further for the first time enables arbitrary style transfers using lightweight MobileNet-based architecture (Fig.~\ref{fig:showcase}(h)).

In sum, our contribution is an innovative dynamic instance normalization (DIN) layer that allows for more efficient and flexible arbitrary style transfers with favorable visual quality in generating challenging style patterns.
This is achieved via a dynamic convolutional transformation
with an elaborate style encoder and a lightweight content encoder.
Experimental results demonstrate that the proposed method yields
results superior to the state of the art both quantitatively and qualitatively,
and meanwhile leads to a significant reduction
of computation cost,
at a factor of twenty.

%\jyc{and meanwhile has more than $20 \times$ complexity reduction}.

%\jyc{yet with more than twenty times faster than prior work.}
%can be summarized as follows: 1) A novel DIN layer that enables a lightweight architecture for arbitrary style transfer with satisfied visual quality and a speedup factor of more than twenty as compared to existing approaches. 2) We show the proposed DIN is a generalized form of existing normalization methods, and show that it provides flexible support for state-of-the-art convolutional operations, which triggers novel functionalities.

% and real-time inference speed on embedded devices.

%\begin{itemize}
%  \item A novel DIN layer that enables a lightweight architecture for arbitrary style transfer with satisfied visual quality and real-time inference speed on embedded devices.
  %enables a flexible and efficient ASPM with fewer computations and better stylization quality.
  %
  %\vspace{-1.3mm}
  %\item A MobileNet-based lightweight network for arbitrary style transfer with state-of-the-art stylization quality.
  %
  %\vspace{-1.3mm}
%  \item We show the proposed DIN is a generalized form of existing normalization methods in theory, and show that it provides flexible support for state-of-the-art convolutional operations, which triggers novel functionalities.
%\end{itemize}

%===============================================================================|=======================================
\section{Related Work}\label{section:RelatedWork}
%===============================================================================|=======================================

%We briefly review here arbitrary style transfer algorithms.

%------------------------------------------------------------------------------

%\noindent\textbf{Arbitrary Style Transfer.}
%
The goal of ASPM style transfer algorithms is to exploit one single trained model to migrate arbitrary artistic styles to a given photo with only one forward pass.
There are two categories of ASPM algorithms in the literature, namely \emph{Non-Parametric ASPM with Markov Random Fields} and \emph{Parametric ASPM with Summary Statistics}.

The idea of non-parametric ASPM is to transfer artistic styles based on local patches.
The first non-parametric ASPM, proposed in \cite{chen2016fast}, divides the content and style activations in the VGG feature space into a set of activation patches, and the target features are then obtained by mapping and swapping each content activation patch with the most similar style patch.
By feeding the target features into the trained decoder, the stylized result can be produced.
Another non-parametric ASPM in \cite{gu2018arbitrary} further adds a constraint upon the algorithm of Chen \textit{et al}., \ie, each patch is required to be mapped only once as possible as it can.
In this way, their algorithm preserves better global style appearance as compared to \cite{chen2016fast}.
Non-parametric ASPM enjoys favorable visual quality but suffers from the heavy computation burden brought by the mapping and swapping procedure.

Parametric ASPM improves the efficiency of non-parametric methods via global summary statistics matching.
Specifically, \cite{huang2017arbitrary} designs a novel \emph{Adaptive Instance Normalization (AdaIN)} layer to explicitly transfer the channel-wise mean and variance statistics between style and content feature activations.
Following their work, \cite{sheng2018neural} further extends AdaIN to multi-scale stylization for better visual quality.

Another line of parametric ASPM is based on \emph{Whitening Instance Normalization (WIN)} proposed by \cite{li2017universal}.
They find that whitening normalization in the VGG feature space can remove the style-related information and meanwhile preserve content structures of input images.
Therefore, they first use whitening instance normalization to filter the style out of the content image, and then apply the coloring transforms to transfer the desired style patterns.
Their proposed WIN-based approach successfully transfers arbitrary styles in a learning free manner. % , as compared with the data-driven AdaIN based methods.
However, their whitening and coloring transforms is realized by matrix computations, which are computationally expensive.
To address this issue, \cite{li2019learning} proposes to learn feed-forward networks to replace the matrix computations in \cite{li2017universal}.

These state-of-the-art ASPM algorithms still suffer from one major flaw: they all require high-cost VGG encoders.
%, which will be further explained in the subsequent section.
%the content image to pass through a pre-trained VGG encoder, which will be further explained in the subsequent section.
%
%This issue limits their usage in mobile and embedded devices.
%
The one exception is the algorithm in \cite{shen2018neural}, which generates a whole 14-layer style-specific stylization network for every style, but leading to expensive cost of extra memory.
%
%However, it has expensive cost of extra memory for every content-style pair.
%
Unlike these existing methods, our approach gets rid of the high-cost encoders and expensive memory, yet with superior quality in challenging styles.
%Unlike these existing methods, our work proposes a novel \emph{Dynamic Instance Normalization} layer for a flexible parametric ASPM with significantly fewer computations and better visual quality.
%arbitrary style transfers, moving towards a more efficient and flexible parametric ASPM with significantly fewer computations and better visual quality.

%------------------------------------------------------------------------------

% Meta Network
%

%-------------------------------------------------------------------------------|---------------------------------------
%\noindent\textbf{Model Compression.}

%-------------------------------------------------------------------------------|---------------------------------------
%\noindent\textbf{Normalization Methods in Style Transfer.}

%XXX
%===============================================================================|=======================================
\section{Proposed Method}\label{section:Method}
%===============================================================================|=======================================

%-------------------------------------------------------------------------------|---------------------------------------
\subsection{Revisiting Normalization Methods in NST}
%-------------------------------------------------------------------------------|---------------------------------------
The development of NST is very related to the emergence of several novel normalization methods.
Here, we revisit and analyze existing normalization methods in the field of NST, which motivate the inspiration of the proposed approach.

\noindent\textbf{\\Normalization for PSPM.}
%\paragraph{PSPM}
The first emerged normalization method in NST, namely \emph{instance normalization (IN)} or \emph{contrast normalization} \cite{ulyanov2016instance}, can be defined as follows:
\begin{equation}
\textrm{IN}(\mathcal{F}_c)= \frac{\mathcal{F}_c-\mu(\mathcal{F}_c)}{\sigma(\mathcal{F}_c)},
\label{eq:IN}
\end{equation}
where $\mathcal{F}_c$ denotes the feature activation given the content image $I_c$ as input.
IN is first and primarily adopted in PSPM algorithms.
It has been demonstrated that compared with \emph{batch normalization}, IN is capable of better visual performance \cite{ulyanov2016instance,ulyanov2017improved} and can also achieve faster and better convergence \cite{huang2017arbitrary}.

\noindent\textbf{\\Normalization for MSPM.}
Built upon the idea of IN, Dumoulin \etal further propose \emph{conditional instance normalization (CIN)} \cite{dumoulin2016learned}, which is to scale and shift the activations in IN layer:
\begin{equation}
\textrm{CIN}(\mathcal{F}_c, \mathcal{F}_s)= \gamma_{\mathcal{F}_s} \times \textrm{IN}(\mathcal{F}_c)+\beta_{\mathcal{F}_s},
\label{eq:CIN}
\end{equation}
where $\mathcal{F}_s$ represents the feature activations of the style image $I_s$, and $\gamma_{\mathcal{F}_s}$ and $\beta_{\mathcal{F}_s}$ are affine parameters corresponding to different styles.
Dumoulin \etal find that by only changing learned $\gamma$ and $\beta$ in CIN layers with different style images, a single network can transfer multiple styles. Their proposed CIN is the earliest MSPM algorithm, and meanwhile achieves state-of-the-art performance even now.
%by learning to tie different pairs of $\gamma$ and $\beta$ to different styles in a data-driven manner, a single network with different values of $\gamma$ and $\beta$ can transfer multiple styles.

\noindent\textbf{\\Normalization for ASPM.} Inspired by the success of CIN, Huang \etal propose to use adaptive affine parameters for arbitrary style transfer, through a novel \emph{adaptive instance normalization (AdaIN)} layer \cite{huang2017arbitrary}.
However, their ``adaptive'' affine parameters are computed in a manually defined manner, \ie, simply using the channel-wise mean $\mu$ and variance $\sigma$ of style features, the most basic statistics, as the affine parameters in AdaIN layer.
Their proposed AdaIN can be formulated as follows:
\begin{equation}
\textrm{AdaIN}(\mathcal{F}_c, \mathcal{F}_s)= \sigma(\mathcal{F}_s)\times \textrm{IN}(\mathcal{F}_c)+  \mu(\mathcal{F}_s).
\label{eq:AdaIN}
\end{equation}

The intuition behind AdaIN is that the mean of VGG feature activations could encode different types of brushstrokes in a certain style, while the variance of the same feature activations could encode the amount of subtle style information, as explained in \cite{huang2017arbitrary}.
%
%Therefore, their proposed AdaIN is based on a relatively strong hypothesis that the mean and variance of style features could totally represent a specific style.
%
%Their intuition is reasonable, leading to an acceptable performance in arbitrary image stylization.
Their algorithm achieves a reasonable performance in arbitrary image stylization.
However, we believe that it is suboptimal to compute these ``adaptive'' affine parameters in such a manually defined way, thus leaving room for improvement.

\begin{figure}[!t]
\setlength\tabcolsep{0.9 pt}
{\renewcommand{\arraystretch}{0.6}
%\begin{tabular}{>{\centering}n{\p} >{\centering}n{\p} >{\centering\arraybackslash}n{\p}}
%\begin{tabular}{>{\centering}m{1.98cm} >{\centering}m{1.98cm} >{\centering}m{1.98cm} ?{0.2mm} >{\centering}m{1.98cm} >{\centering}m{1.98cm} >{\centering\arraybackslash}m{1.98cm}}
\begin{tabular}{cccc}
\centering

\includegraphics[width=0.115\textwidth]{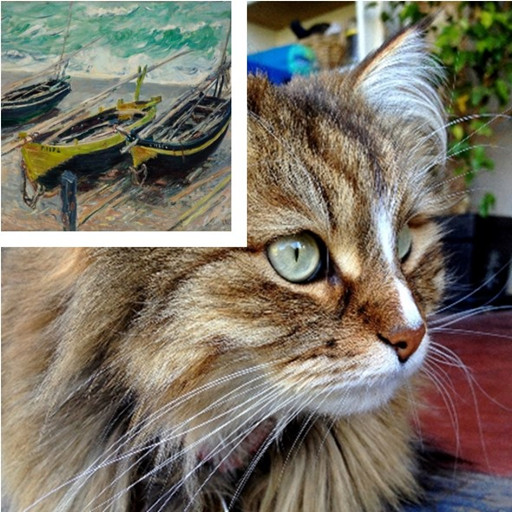}& \includegraphics[width=0.115\textwidth]{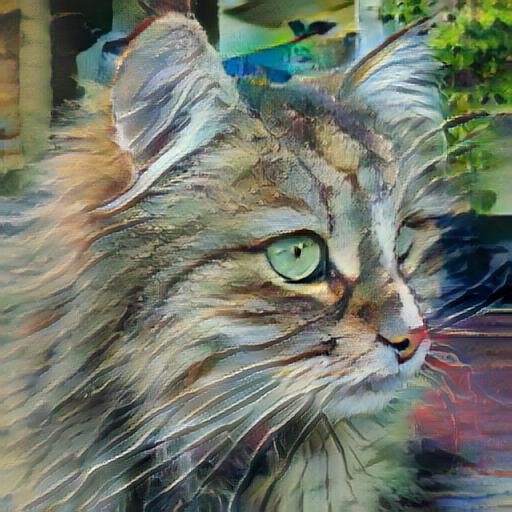} & \includegraphics[width=0.115\textwidth]{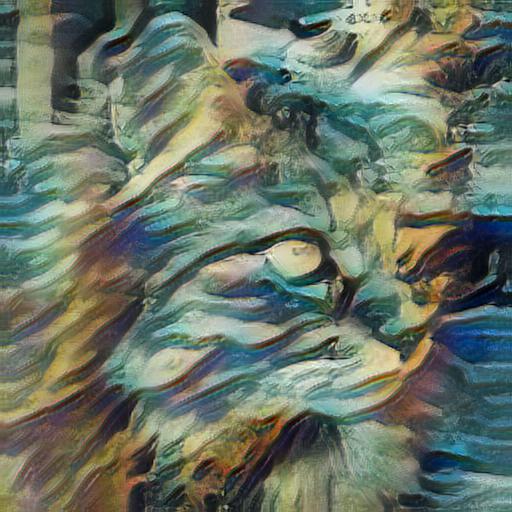} &\includegraphics[width=0.115\textwidth]{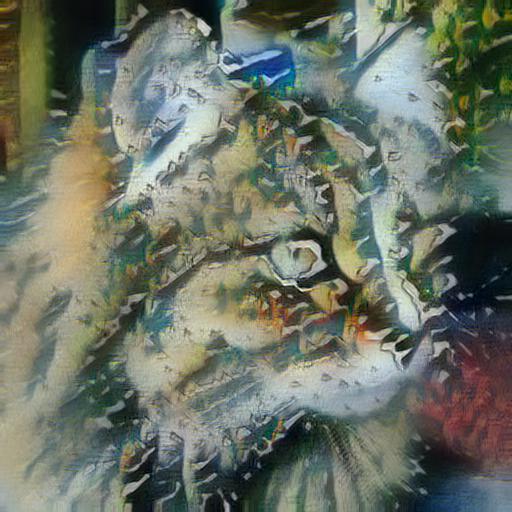}\\
\includegraphics[width=0.115\textwidth]{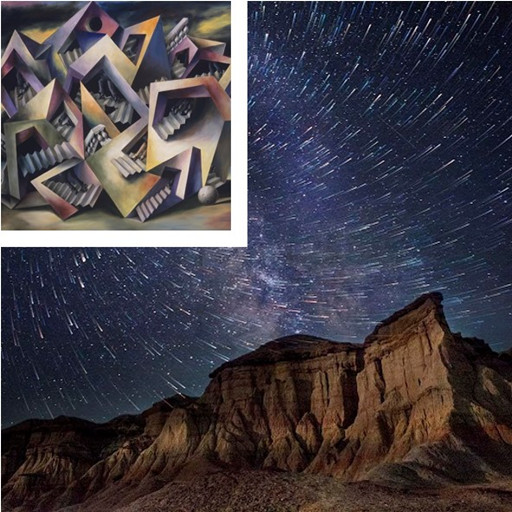}& \includegraphics[width=0.115\textwidth]{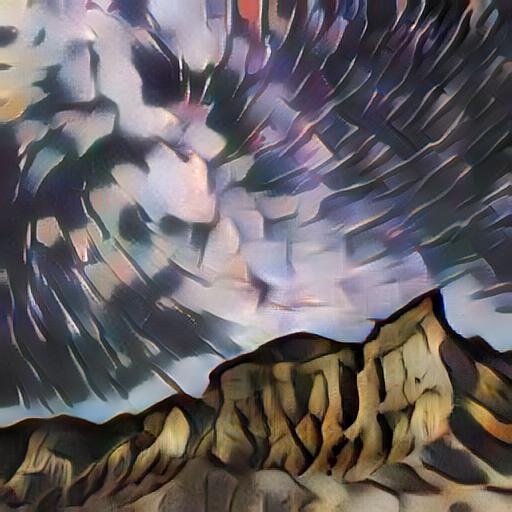} & \includegraphics[width=0.115\textwidth]{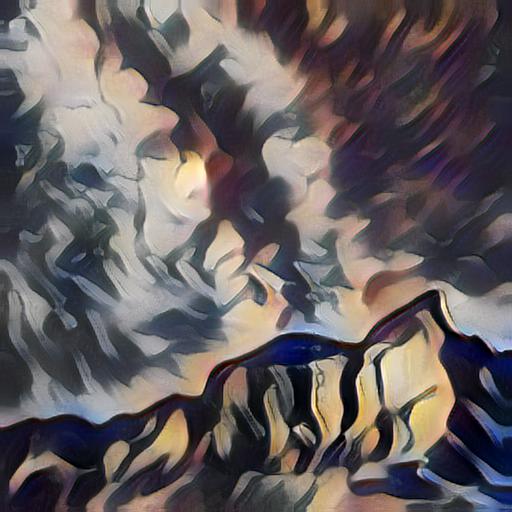} &\includegraphics[width=0.115\textwidth]{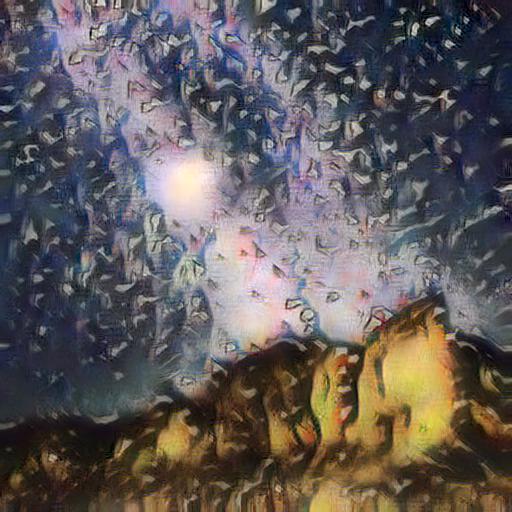}\vspace{0.1cm}\\
%\smallskip
\scriptsize{Content \& Style}&\scriptsize{Shared VGG \emph{Enc}} & \scriptsize{Different VGG \emph{Enc}} & \scriptsize{Shared Smaller \emph{Enc}}
%\scriptsize{(a)}&\scriptsize{(b)} & \scriptsize{(c)} & \scriptsize{(d)}
\end{tabular}
}
\vspace{-0.3cm}
%\vspace{-0.01cm}
\caption{Stylization results of using shared VGG encoders (\nth{2} column), different VGG encoders (\nth{3} column) and shared but much smaller encoders (\nth{4} column) for producing separate content and style features in AdaIN layer. \emph{Enc} represents encoders.}
\label{fig:analysis} %% label for entire figure
\end{figure}

More importantly, we find two implicit requirements for the favorable performance of AdaIN in arbitrary stylization:
\noindent\textbf{1)} The content and style encoders for producing $\mathcal{F}_c$ and $\mathcal{F}_s$ in Eq.~\ref{eq:AdaIN} should be kept the same;\\ \noindent\textbf{2)} The network architecture of the encoders should be complex enough to extract high-level encoded features, like the VGG network.% used in \cite{huang2017arbitrary}.

To validate our first point, we show the comparison results of using shared VGG encoders (Fig.~\ref{fig:analysis}, \nth{2} column) and using different VGG encoders (Fig.~\ref{fig:analysis}, \nth{3} column) for producing $\mathcal{F}_c$ and $\mathcal{F}_s$.
%
%Fig.~\ref{fig:analysis} validates our first point by demonstrating the comparison results of using the same VGG encoders (\nth{2} column) and using different VGG encoders (\nth{3} column).
%
The results of using different VGG encoders are generated by setting the original style encoder in \cite{huang2017arbitrary} as trainable while the content VGG encoder remains unchanged.
Other experimental settings remain the same. % for the results in Fig.~\ref{fig:analysis}.
The results with different encoders (Fig.~\ref{fig:analysis}, \nth{3} column), as can be observed, are inferior in visual quality with unexpected patterns, primarily because $\mathcal{F}_c$ and $\mathcal{F}_s$ are no longer in the same feature space.

Also, in the \nth{4} column of Fig.~\ref{fig:analysis}, we try to use a smaller network architecture \cite{Johnson2016perceptual} to replace the original VGG encoders of AdaIN.
In particular, the content and style encoders are shared, like the \nth{2} column in Fig.~\ref{fig:analysis}.
%when producing $\mathcal{F}_c$ and $\mathcal{F}_s$,
The visual results of smaller encoders, as can be observed, are less appealing with many artifacts, which is consistent with our second point.
We believe that the reason is: the way of manually defining the affine parameters as feature mean and variance in AdaIN needs a more meaningful high-level feature representation, which requires the encoder to be deep enough.
This point is also partly observed in \cite{shen2018neural}.

These two implicit requirements reveal another major flaw of AdaIN: neither of the content and style encoder can be reduced to a lightweight one, which makes the network cumbersome for deployment in resource-constrained environments.
\emph{Whitening Instance Normalization (WIN)}, another group of ASPM, also has this issue, since whitening and coloring transforms also need meaningful features, as explained in related work.
% caption: The fixed encoder for the style image; the trainable encoder for the content image.

%The other group of methods based on whitening instance normalization layer also suffers from this as mentioned in Section~\ref{section:RelatedWork}.

%\begin{equation}
%\textrm{IN}(\mathcal{F}(I_c))= \frac{\mathcal{F}(I_c)-\mu(\mathcal{F}(I_c))}{\sigma(\mathcal{F}(I_c))},
%\label{eq:IN}
%\end{equation}
%
%\begin{equation}
%\textrm{CIN}(\mathcal{F}(I_c), \mathcal{F}(I_s))= \gamma_{\mathcal{F}(I_s)} \times \textrm{IN}(\mathcal{F}(I_c))+\beta_{\mathcal{F}(I_s)},
%\label{eq:CIN}
%\end{equation}
%
%\begin{equation}
%\textrm{AdaIN}(\mathcal{F}(I_c), \mathcal{F}(I_s))= \sigma(\mathcal{F}(I_s))\times \textrm{IN}(\mathcal{F}(I_c))+  \mu(\mathcal{F}(I_s)).
%\label{eq:AdaIN}
%\end{equation}
%
%\begin{equation}
%\textrm{DIN}(\mathcal{F}(I_c), \mathcal{F}(I_s))= f[\mathcal{F}(I_s), \textrm{IN}(\mathcal{F}(I_c))]
%\label{eq:AdaIN}
%\end{equation}

%-------------------------------------------------------------------------------|---------------------------------------
%\subsection{Generalized Normalization Model for NST}
%-------------------------------------------------------------------------------|---------------------------------------
\begin{figure}[!t]
  \centering
  \includegraphics[width=0.38\textwidth]{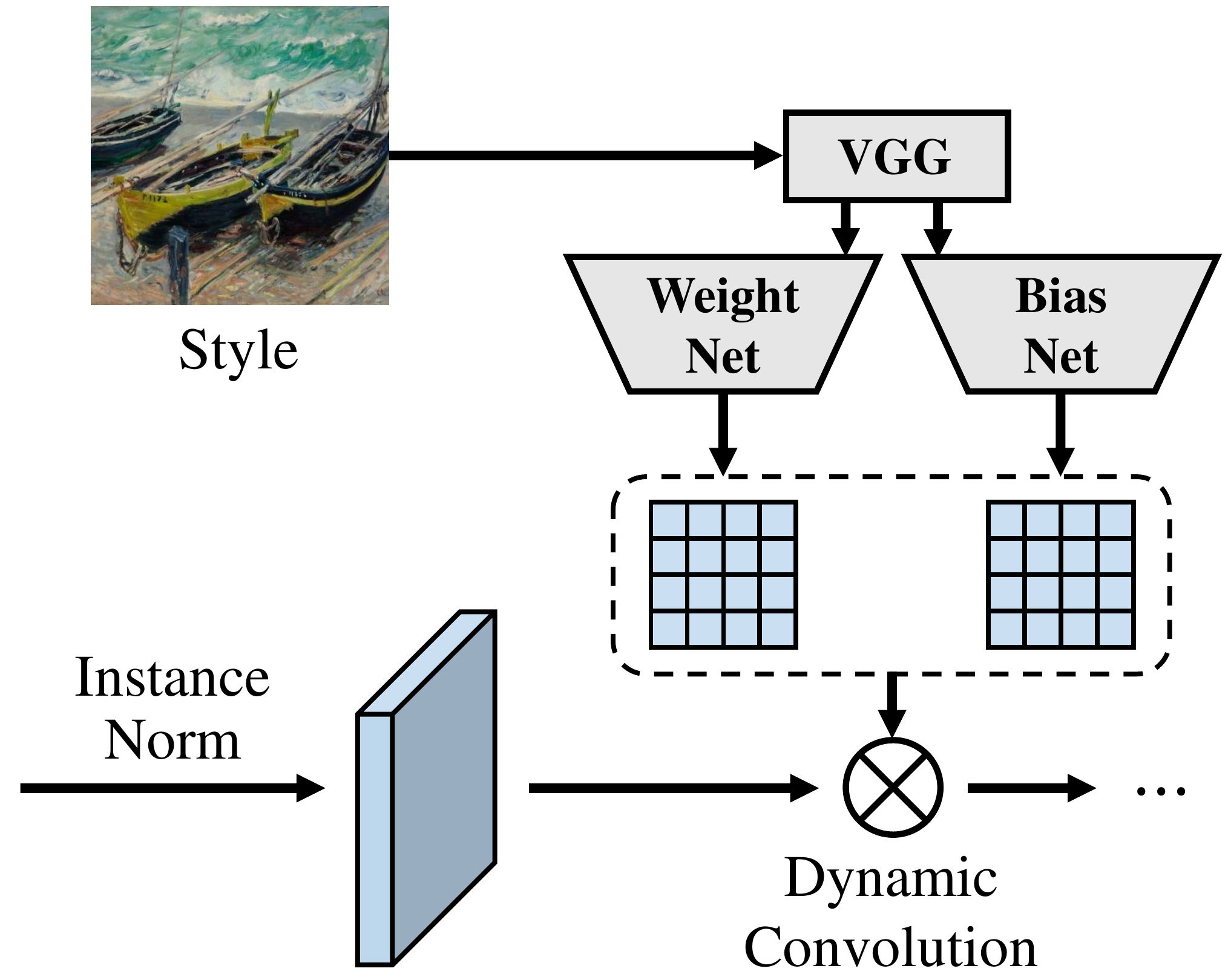}\\
  \caption{DIN layer consists of an instance normalization and a dynamic convolutional operation.
  Here, the convolution types include
  but are not limited to
  standard convolution, deformable convolution, and
  groupwise convolution.}\label{fig:layer}
\end{figure}

%-------------------------------------------------------------------------------|---------------------------------------
\subsection{Dynamic Instance Normalization}
%-------------------------------------------------------------------------------|---------------------------------------
To address the aforementioned limitations in existing ASPM algorithms, we propose a new and generalized \emph{Dynamic Instance Normalization (DIN)} layer for arbitrary image stylization.
Instead of manually defining the way to compute the affine parameters so as to align the mean and variance, the simplest statistics, between content and style features for arbitrary style transfer, we introduce a more generalized dynamic convolutional transformation, of which the parameters are adaptively changed in a learnable manner according to different styles, leading to a more accurate alignment of the real complex statistics of style features.
%according to different input styles

Given a pair of content image $I_c$ and style image $I_s$ as input, the proposed DIN layer can be modelled as:
\begin{equation}
\textrm{DIN}(\mathcal{F}_c, \mathcal{F}_s)= f[\mathcal{F}_s, \textrm{IN}(\mathcal{F}_c)],
\label{eq:DIN}
\end{equation}
where $\mathcal{F}_c$ and $\mathcal{F}_s$ are the corresponding feature representations of $I_c$ and $I_s$, and $f$ is a dynamic convolutional operation \cite{jia2016dynamic}.
Unlike standard convolutions of which the weight and bias are model parameters, in our dynamic convolution $f$, the weight and bias are dynamically generated by encoding different input style images.
%as the feature maps of inputting $\mathcal{F}_s$ into the corresponding weight and bias generators.

Fig.~\ref{fig:layer} illustrates the basic structure of the proposed DIN, comprising an instance normalization, like previous CIN (Eq.~\ref{eq:CIN}) and AdaIN (Eq.~\ref{eq:AdaIN}), and a dynamic convolution, of which the parameters are adaptively changed by forwarding different input styles to separate weight and bias networks.
In this way, the proposed DIN can be regarded to encode a given style image into learnable convolutional parameters via a sophisticated style encoder, and the content image is then stylized by the dynamic convolutional transformation.
%
%The DIN layer is comprised of two components. The first component is an IN layer, like previous normalization methods in Eq.~\ref{eq:CIN} and Eq.~\ref{eq:AdaIN}.
%
%Following the IN component, a dynamic convolutional layer is introduced, which takes the output of IN as input and dynamically changes its parameters (\ie, weight and bias) according to the given style image.
%
%The weight and bias of the dynamic convolution are generated with two separate weight and bias networks, which are composed of two convolutional and adaptive pooling operations for handling input style images of arbitrary size.
The weight and bias networks consist of two convolutional and adaptive pooling operations for handling arbitrary input sizes, of which the output size is set according to the type and hyperparameters of the dynamic convolution.
%

%the output size of the weight and bias networks would need to be adjusted according to the kernel size of the dynamic convolutional operation.
%
%The output feature size of the weight and bias networks would need to be adjusted according to the kernel size of the dynamic convolutional operation.

%In sum, unlike prior normalization methods that manually define the way to calculate affine parameters, our proposed DIN changes the previous affine transformation to a more generalized and powerful dynamic convolutional transformation, of which the parameters are learned to be adaptively changed for a better match of the real distributions of complex styles.
%
%The parameters of the dynamic convolution are learned to be adaptively changed for a better match of the real distributions of complex styles.

Compared with prior normalization methods, first, the proposed DIN model in Eq.~\ref{eq:DIN} is much more generalized, including all the aforementioned existing normalization methods as special cases.
Specifically, IN in Eq.~\ref{eq:IN} can be treated as a special case of our model in Eq.~\ref{eq:DIN} where the weight and bias of $f$ are set to $0$ and $1$, respectively.
CIN in Eq.~\ref{eq:IN} corresponds to the case where both of the weight and bias in $f$ are scalars.
AdaIN in Eq.~\ref{eq:AdaIN} is also a special case of the proposed DIN when the channel-wise variance and mean of style features are manually set as the weight and bias of $f$.
%
%\jyc{Therefore, the proposed DIN is a generalized form of existing normalization methods, leading to a larger search space for better optimized solutions and convergence.}
The proposed DIN therefore can be treated as a generalized framework to normalization, under which existing normalization methods are taken as specific realizations, allowing for a larger search space for better optimized solutions and convergence.

%Even with the same settings as previous normalization methods, this generalized property of our proposed DIN could lead to at least equal or superior performance.

Second, and more importantly, the proposed DIN does not have the aforementioned requirement of complex and shared encoders in previous ASPM methods, since we do not need to align the mean and variance of meaningful feature activations between content and style for stylization, thanks to our dynamic convolutional transformation.
By contrast, the proposed approach enables a more sophisticated style encoder with dynamic convolution, so as to encode rich and complex style patterns adequately, yet using a more lightweight content encoder for faster inference, since the style-specific parameters for known styles can be stored in advance.
As a result, the proposed DIN yields superior results in transferring challenging style patterns and fine strokes, and meanwhile conduct arbitrary stylizations with an over $20 \times$ reduction in computation costs, as compared with the state of the art.
%achieves a speedup factor of more than twenty for inference.

%By contrast, the proposed approach makes it possible to use smaller encoders for both content and style images, or use a much smaller content encoder but a relatively larger style encoder (\ie, moving the computational burden to encode style information since we can use stored style-specific parameters for known styles).
%

Third, the proposed DIN provides flexible support for a variety of convolutional operations, by simply changing the type of the dynamic convolution in Fig.~\ref{fig:layer}.
In particular, by incorporating some state-of-the-art convolutional operations, the proposed DIN is able to create novel functionalities in stylization.
Here, we explain two variants of the proposed DIN as examples, which are deformable DIN and spatially-adaptive DIN.
%is more flexible, since our dynamic convolutional operation $f$ in Fig.~\ref{fig:layer} supports all kinds of state-of-the-art convolutional operations other than linear operations in previous normalization methods.
%
%In particular, novel functionalities that are not possible before can be triggered by incorporating specific convolutional operations into our proposed DIN:
%
\\ \noindent\textbf{1) Deformable Dynamic Instance Normalization.}
%\paragraph{(1) Deformable Dynamic Instance Normalization.}
%
By using the deformable convolutional operation \cite{dai2017deformable} for $f$, our algorithm achieves the first automatic spatial-stroke control in arbitrary style transfers.
In deformable convolutions, the receptive fields and sampling locations can be adaptively adjusted according to the foreground objects' shape and scale.
The deformable dynamic convolutional kernel, therefore, obtains the ability to stylize images in an attention-aware manner.
For foreground and background objects, the deformable dynamic convolutional kernel would use different strokes according to the visual attention.
As compared to existing approaches that randomly place different strokes across the whole image, the achieved automatic spatial-stroke control makes AI-created art much closer to human-created art.
\\ \noindent\textbf{2) Spatially-Adaptive Dynamic Instance Normalization.}
Prior ASPM algorithms based on IN suffer from another issue: They cannot generate proper strokes for uniform pixel areas of the input content image \cite{huang2018multimodal}.
The reason is that the convolutional output of the uniform pixel areas also has uniform values.
After normalizing these uniform values through the widely adopted IN layer, the output activation would become all zeros, thereby preventing the proper stroke generations in the corresponding area.
This limitation is especially serious for hand-crafted non-natural content images, where the uniform pixel areas are quite common.
The issue can be addressed by setting the kernel size of $f$ in the proposed DIN as the input feature map size, resulting in a special spatially-adaptive convolution similar to the operation used in \cite{park2019semantic}.
%increasing the kernel size of $f$ in our DIN to the input feature map size, resulting in a special spatially-adaptive convolution similar to the operation in \cite{park2019semantic}.
%
%In this way, our dynamic convolutional operation in DIN would evolve to a special spatially-adaptive convolution (to some extend similar to the operation in \cite{park2019semantic}).
%
In this way, the activation values for uniform pixel areas would not be uniform, leading to proper stroke generations in corresponding areas.

\begin{figure*}[!t]
  \centering
  \includegraphics[width=0.85\textwidth]{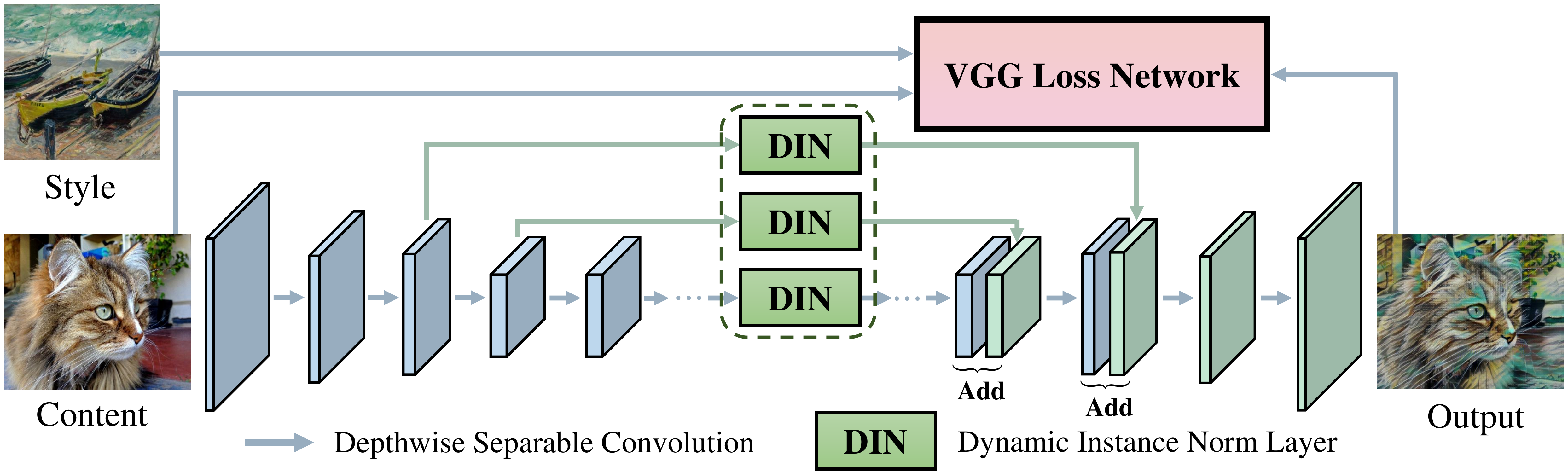}\\
  \caption{Network architecture of the proposed hierarchical lightweight arbitrary style transfer network based on MobileNetV1. DIN represents the proposed dynamic instance normalization layer as depicted in Fig.~\ref{fig:layer}.}\label{fig:arch}
\end{figure*}

%-------------------------------------------------------------------------------|---------------------------------------
\subsection{An Application of Dynamic Instance Normalization}
%-------------------------------------------------------------------------------|---------------------------------------
%
We introduce here one major application of the proposed DIN in lightweight arbitrary style transfers for deployment in mobile and embedded devices.
In particular, with the proposed DIN layer, we build the first MobileNet-based lightweight arbitrary stylization network in the literature, based on depthwise separable convolutions in MobileNetV1 \cite{howard2017mobilenets}.

Fig.~\ref{fig:arch} shows the proposed MobileNet-based network architecture, which mainly consists of three modules: image encoder, dynamic instance normalization layer and image decoder.
The first module, image encoder, comprises one standard stride-1 convolutional layer, two stride-2 depthwise separable convolutional blocks and two stride-1 residual layers, inspired by the network design in \cite{Johnson2016perceptual}.
%
%More specifically, each depthwise separable convolutional block contains a depthwise convolution and a pointwise convolution.
%
The image decoder is symmetric with the encoder, but with the pooling layers replaced by bilinear upsampling layers.
In particular, the intermediate feature maps of the image decoder receive hierarchical normalized features from several separate DIN layers for additions.

The idea of this hierarchical DIN design comes from the traditional image processing algorithms, where images are generally decomposed hierarchically and each hierarchical level contains image information of a specific level.
Since the features from a deep network also have a similar hierarchical structure, we follow and apply this idea in our network design.
By hierarchically normalizing different levels of feature representations, the proposed DIN layers introduce multi-level style information for arbitrary style transfers.
Although the single-level stylization with one single DIN layer also works in our experiment, the proposed hierarchical multi-level stylization can generate different scales of style elements better, and meanwhile preserve more content structures.
%As a result, compared with single-level stylization, our multi-level network could generate different scales of style elements better.
%
More detailed architecture designs of the proposed network can be found in the supplementary material.

%%-------------------------------------------------------------------------------|---------------------------------------
%\paragraph{Loss Functions}
%%-------------------------------------------------------------------------------|---------------------------------------
%
%
%The semantic loss is defined to preserve the semantic information in the content image, which is formulated as the Euclidean distance between the content image $I_c$ and the output stylized image $I_o$ in the feature space of the VGG network \cite{gatys2016image}.
%
%Assume that $\mathcal{F}^{l}(I) \in \mathbb{R}^{C \times H \times W}$ represents the feature map at layer $l$ in VGG network with a given image $I$, where $C$, $H$ and $W$ denote the number of channels, the height and width of the feature map respectively. The semantic content loss is then defined as:
%
%\begin{equation}
%\mathcal{L}_{c} =  \sum\mathop{}_{l \in \{l_c\}}\lVert \mathcal{F}^{l}(I_c) - \mathcal{F}^{l}(I_o) \rVert^2,
%\end{equation}
%where $\{l_c\}$ represents the set of VGG layers used to compute the content loss.
%
%
%Our loss functions have two components. The first component is the content loss, which is defined as the Eclidean distance between the output image and the content image in the VGG feature space.
%
%
%The style loss is defined as the variance and mean
%
%in ablation studies, we will  different settings
%in our following sections

%===============================================================================|=======================================
\section{Experiments}\label{section:Experiments}
%===============================================================================|=======================================
%-------------------------------------------------------------------------------|---------------------------------------
\subsection{Implementation Details}
%-------------------------------------------------------------------------------|---------------------------------------
By default, the filter size of the proposed DIN layer is set to $1 \times 1$, considering the computational cost.
We use the perceptual loss proposed in \cite{Johnson2016perceptual} as our content loss, and the BN-statistic loss proposed in \cite{li2017demystifying} as our style loss, which are widely adopted in NST.
%
%Our total loss is a weighted sum of these two loss terms.
%
Similar to prior works \cite{gatys2016image,huang2017arbitrary,jing2018stroke}, we use a pre-trained VGG-19 as our loss network.
The content loss is computed at layer $\{relu4\_1\}$, while the style loss is computed at layer $\{ relu1\_1, relu2\_1, relu3\_1, relu4\_1 \}$ of the VGG network.
During training, we adopt the Adam optimizer \cite{kingma2014adam}.
The learning rates for both the image encoder and decoder are set to $0.0001$.
The weight and bias networks in DIN layers are set to have a $10 \times$ learning rate for faster convergence.
%
%The batch is set to $8$.
%
%Our implementation is based on PyTorch \cite{paszke2017automatic}.
%
The training takes roughly one day on an NVIDIA Tesla V100 GPU.
%
%More information can be found in the supplementary material.

\newcommand\z{2.15cm}
\newcommand\x{0.122}
\begin{figure*}[!t]
\setlength\tabcolsep{1 pt}
{\renewcommand{\arraystretch}{0.46}
%\begin{tabular}{>{\centering}n{\p} >{\centering}n{\p} >{\centering\arraybackslash}n{\p}}
%\begin{tabular}{>{\centering}m{1.98cm} >{\centering}m{1.98cm} >{\centering}m{1.98cm} ?{0.2mm} >{\centering}m{1.98cm} >{\centering}m{1.98cm} >{\centering\arraybackslash}m{1.98cm}}
%\begin{tabular}{cccccccc}
\begin{tabular}{>{\centering}m{\z} >{\centering}m{\z} >{\centering}m{\z} >{\centering}m{\z} >{\centering}m{\z} >{\centering}m{\z} >{\centering}m{\z} >{\centering\arraybackslash}m{\z}}
\centering

\includegraphics[width=\x\textwidth]{figs/Quality/style4/style_img.jpg}&
\includegraphics[width=\x\textwidth]{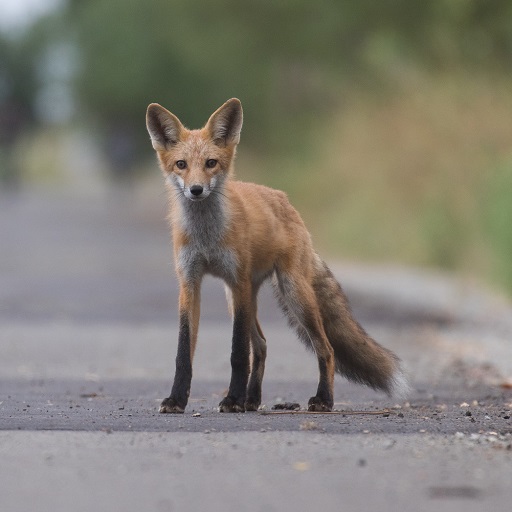}&
\includegraphics[width=\x\textwidth]{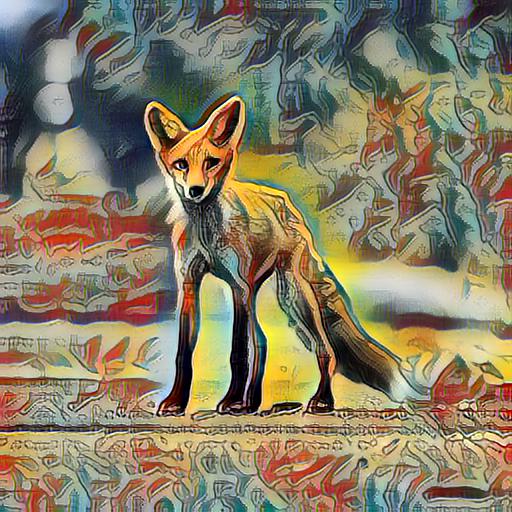}&
\includegraphics[width=\x\textwidth]{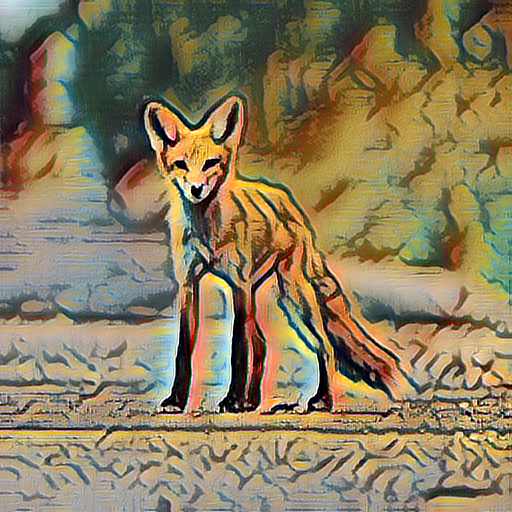}&
\includegraphics[width=\x\textwidth]{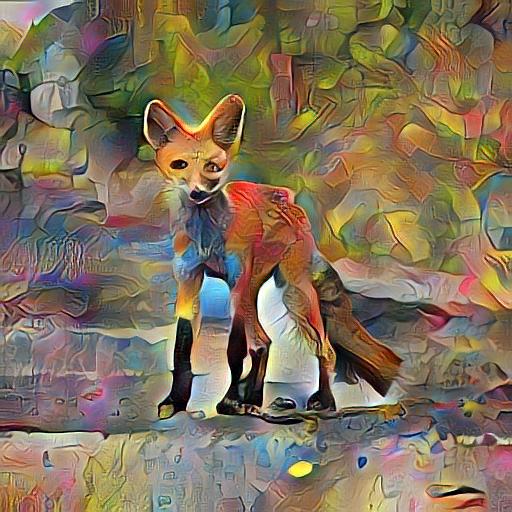}&
\includegraphics[width=\x\textwidth]{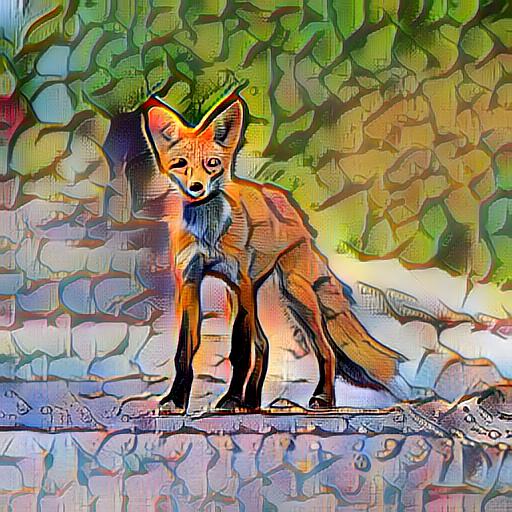}&
\includegraphics[width=\x\textwidth]{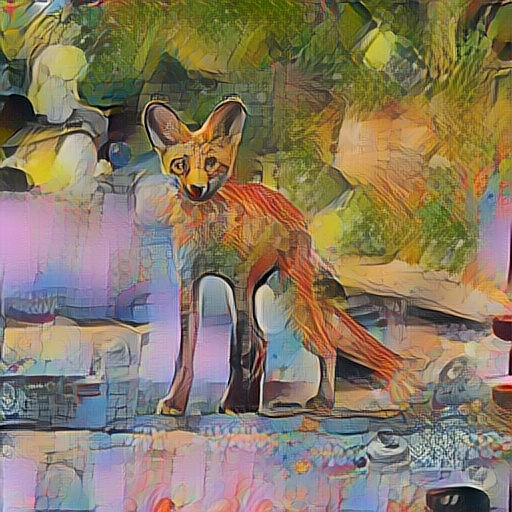}&
\includegraphics[width=\x\textwidth]{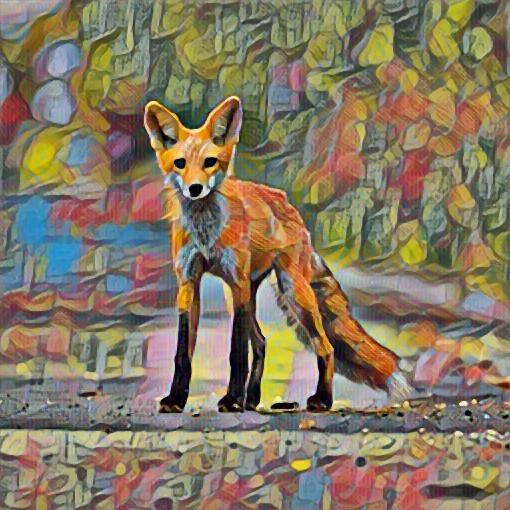}\\

\includegraphics[width=\x\textwidth]{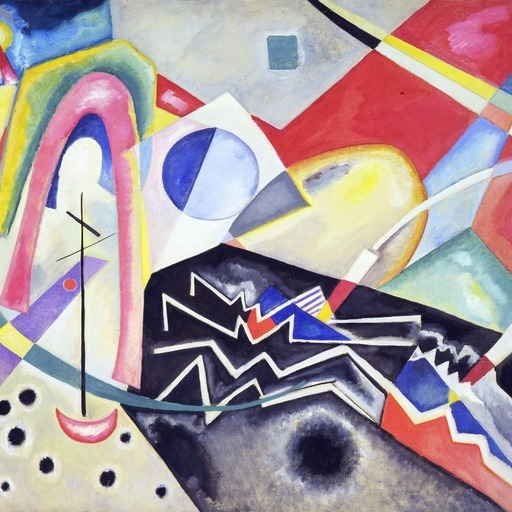}&
\includegraphics[width=\x\textwidth]{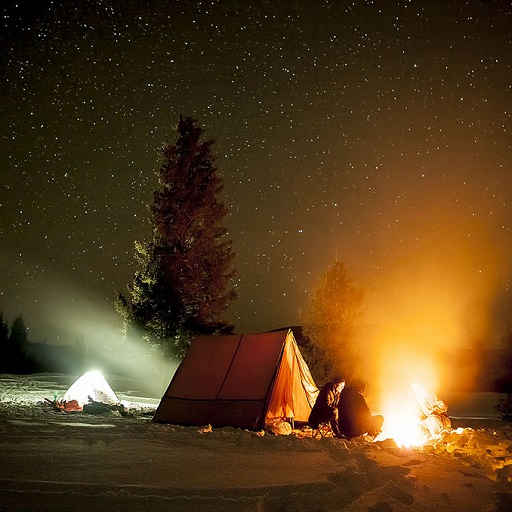}&
\includegraphics[width=\x\textwidth]{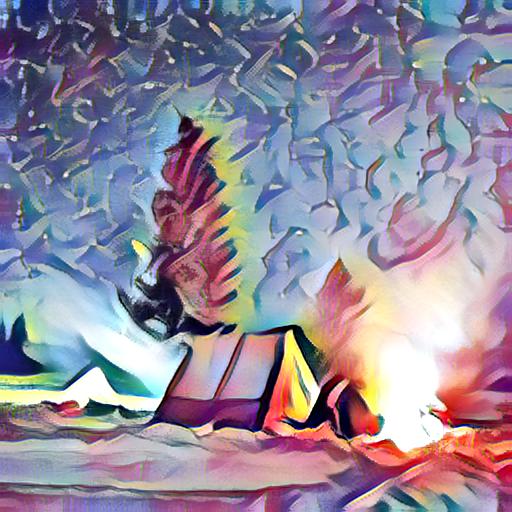}&
\includegraphics[width=\x\textwidth]{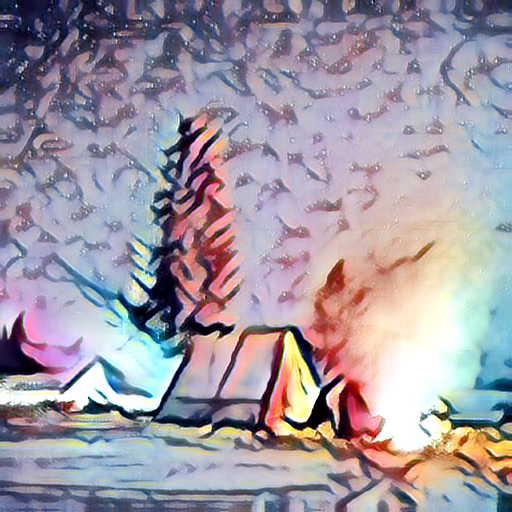}&
\includegraphics[width=\x\textwidth]{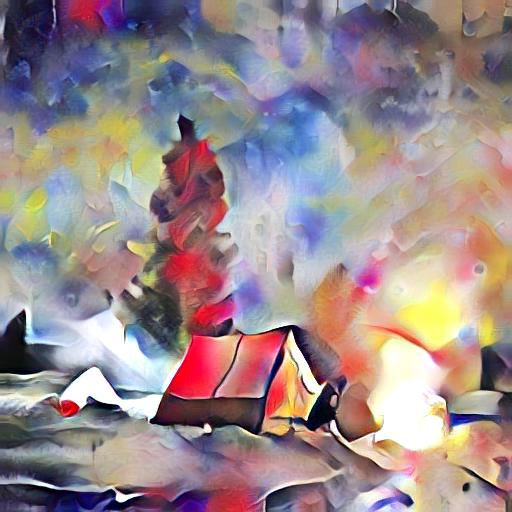}&
\includegraphics[width=\x\textwidth]{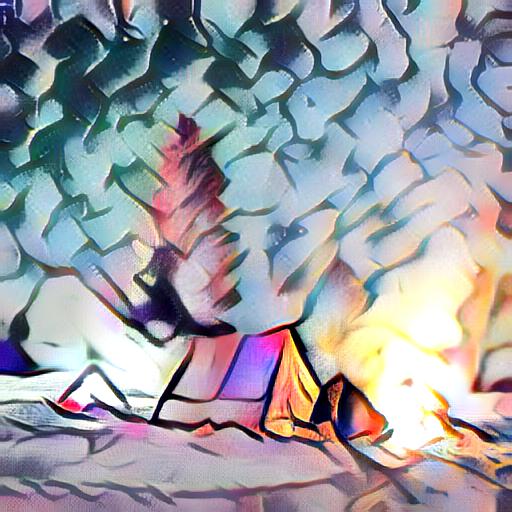}&
\includegraphics[width=\x\textwidth]{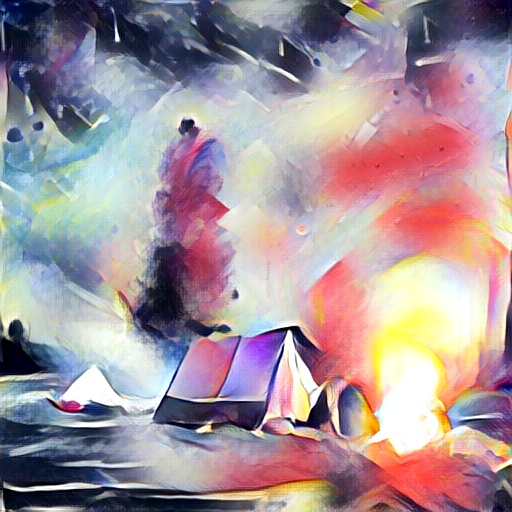}&
\includegraphics[width=\x\textwidth]{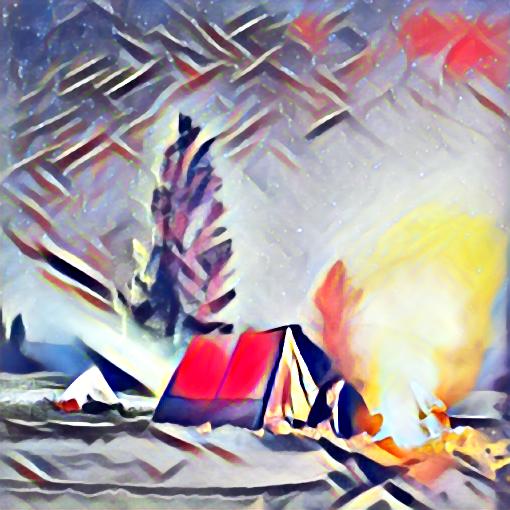}\\

\includegraphics[width=\x\textwidth]{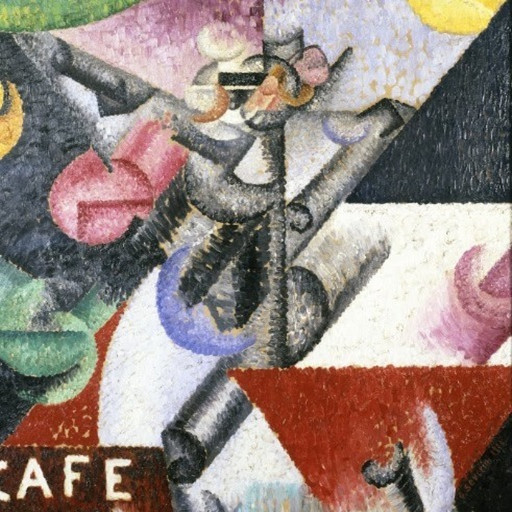}&
\includegraphics[width=\x\textwidth]{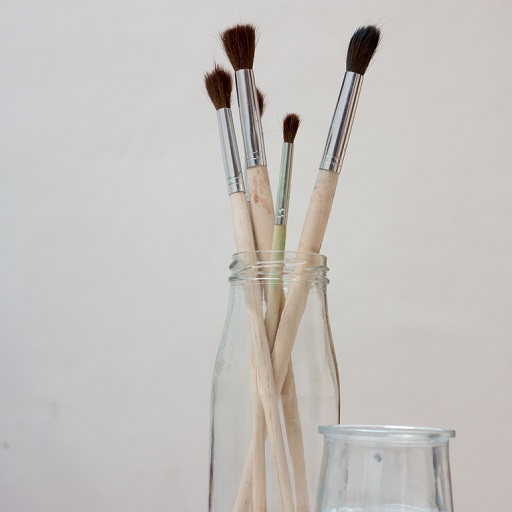}&
\includegraphics[width=\x\textwidth]{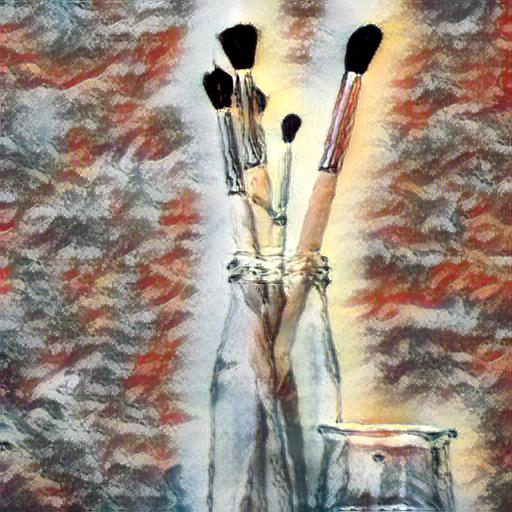}&
\includegraphics[width=\x\textwidth]{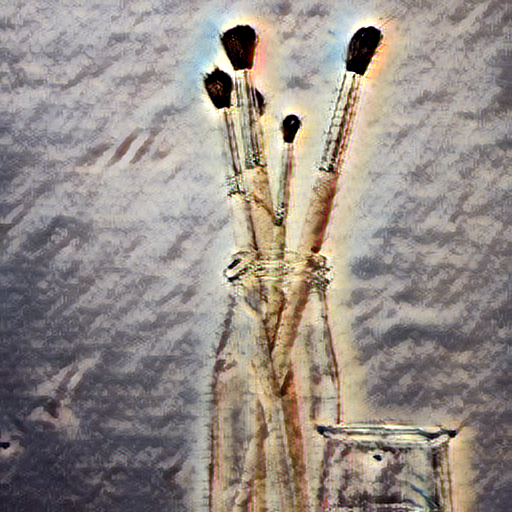}&
\includegraphics[width=\x\textwidth]{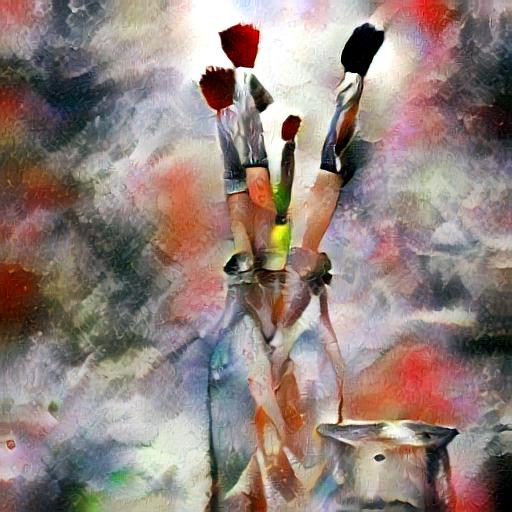}&
\includegraphics[width=\x\textwidth]{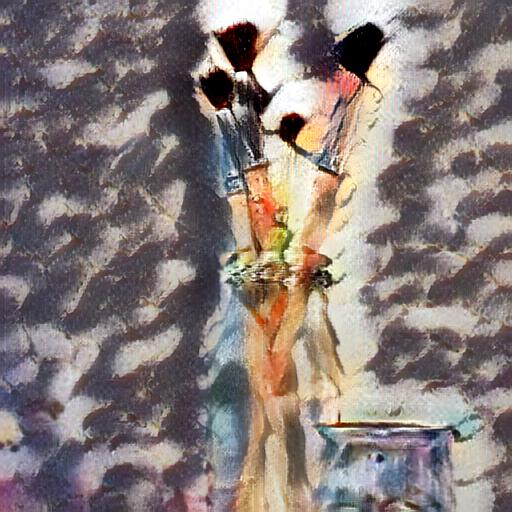}&
\includegraphics[width=\x\textwidth]{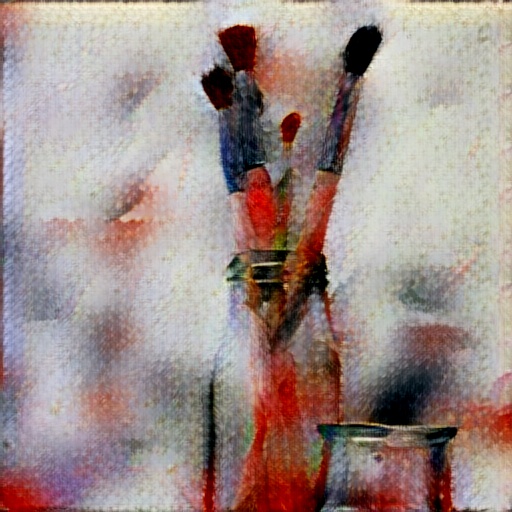}&
\includegraphics[width=\x\textwidth]{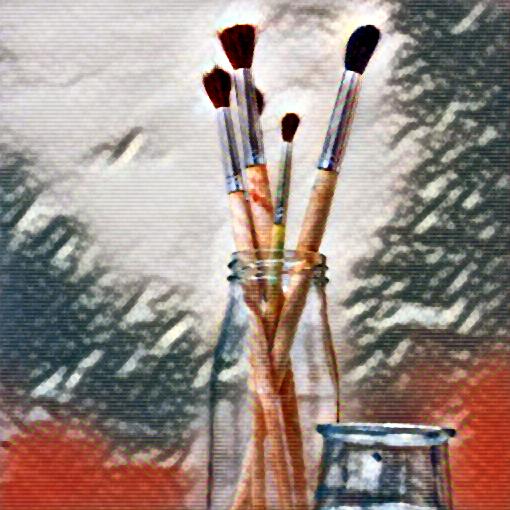}\\

\includegraphics[width=\x\textwidth]{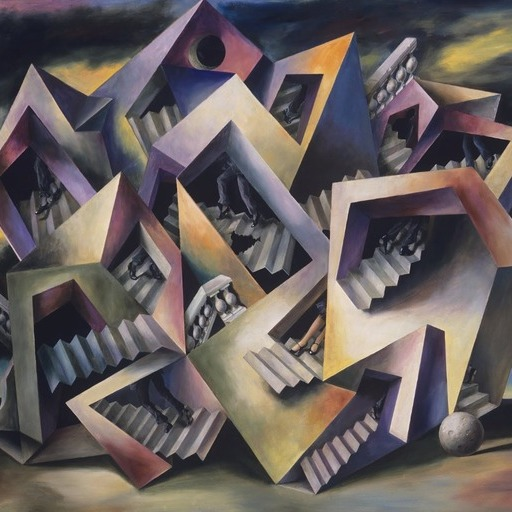}&
\includegraphics[width=\x\textwidth]{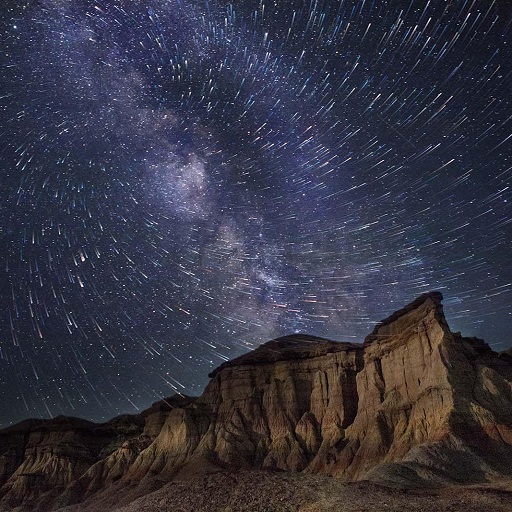}&
\includegraphics[width=\x\textwidth]{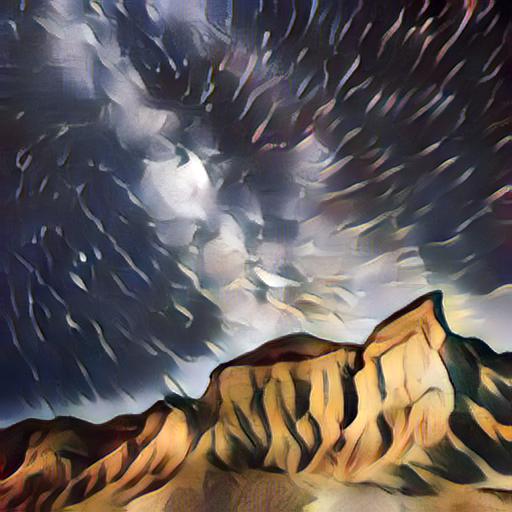}&
\includegraphics[width=\x\textwidth]{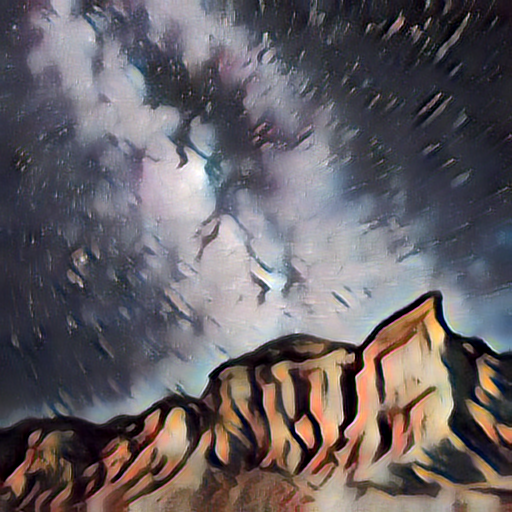}&
\includegraphics[width=\x\textwidth]{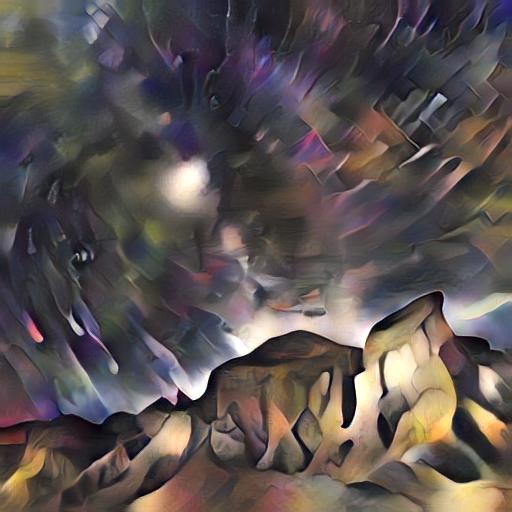}&
\includegraphics[width=\x\textwidth]{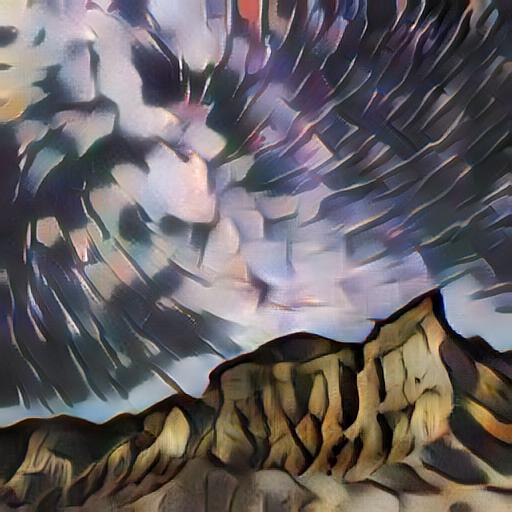}&
\includegraphics[width=\x\textwidth]{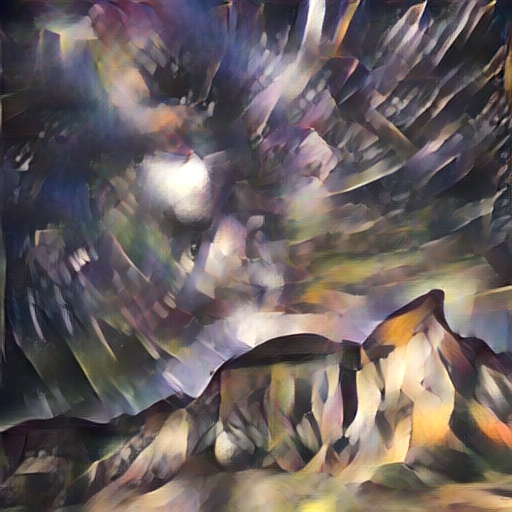}&
\includegraphics[width=\x\textwidth]{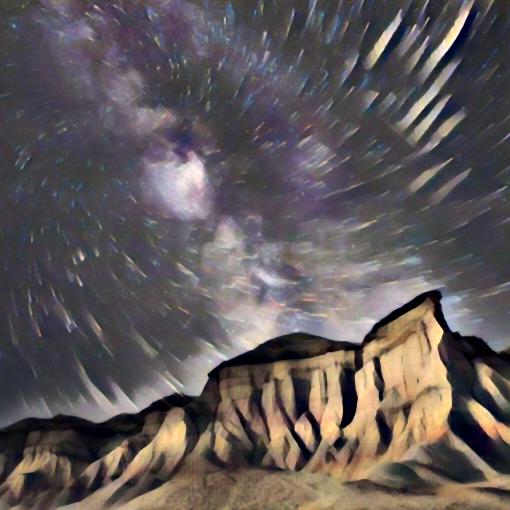}\\

\vspace{0.08cm}\small{Style} & \vspace{0.08cm}\small{Content} & \vspace{0.08cm}\small{Ours~(VGG)}& \vspace{-0.128cm} \multirow{1}{2.2cm}{\small{Ours~(MobileNet)}} & \vspace{0.08cm}\small{Li \etal}& \vspace{0.08cm}\small{Huang \etal} & \vspace{0.08cm}\small{Sheng \etal}& \vspace{0.08cm}\small{Johnson \etal}\\

%\small{\vspace{0.1cm}Content \& Style} & \small{\vspace{0.1cm}Ours (VGG)}& \small{\vspace{0.1cm}Ours (MobileNet)} & \small{\vspace{0.1cm}Li \etal}& \multirow{1}{2.7cm}{\small{Huang \etal}} & \small{\vspace{0.1cm}Sheng \etal}& \small{\vspace{0.1cm}Johnson \etal}
%\scriptsize{(a)}&\scriptsize{(b)} & \scriptsize{(c)} & \scriptsize{(d)}
\end{tabular}
}
\vspace{-0.2cm}
%\vspace{-0.01cm}
\caption{Qualitative  results of our proposed ASPM stylization algorithm and other methods. All the testing content and style images are not used in training. Addition experimental results can be found in the supplementary material.}
\label{fig:quality} %% label for entire figure
\end{figure*}

%-------------------------------------------------------------------------------|---------------------------------------
\subsection{Datasets}
%-------------------------------------------------------------------------------|---------------------------------------
Our network is trained on $82,783$ content images from Microsoft COCO dataset \cite{lin2014microsoft}, and $79,433$ style images from WikiArt \cite{wikiart}.
%
%Before training, all the images are cropped and resized to $512 \times 512$ pixels.
%
%For inference, in the current community of NST, it is common that researchers randomly select some style and content images for comparison.
%
%We believe that this is suboptimal, since the diversity of testing images is not guaranteed.
%
%Therefore, in our experiments, we try to build a testing dataset that contains diversified content and style images.
For inference, since there are no standard datasets in the current field of NST, we try to build here a new public testing dataset that contains diversified content and style images for evaluations.
Specifically, we collect forty content images from \emph{flickr.com}, containing roughly equivalent numbers of four categories: still life photos, portrait photos, animal photos, and landscape photos.
For the style images, we select eight styles from \emph{Google Arts \& Culture}, including abstract, cubism, impressionism, surrealism, futurism, contemporary and expressionism.
All the testing images are not used in training.
The proposed testing dataset can be found in the supplementary material, which will also be publicly available.

\begin{figure}[!t]
\setlength\tabcolsep{0.8 pt}
{\renewcommand{\arraystretch}{0.6}
\begin{tabular}{ccc}
\centering
%\vspace{-0.02cm}

\includegraphics[width=0.155\textwidth]{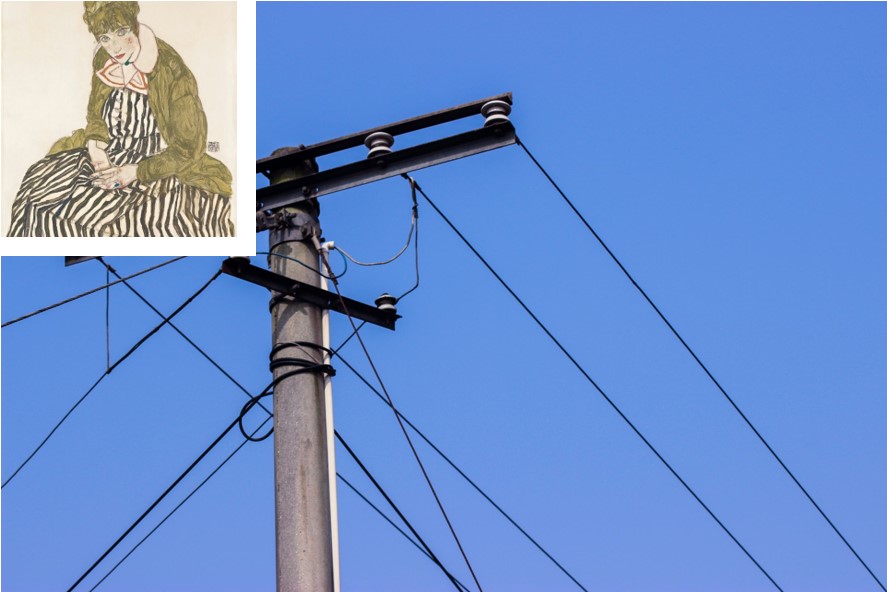}& \includegraphics[width=0.155\textwidth]{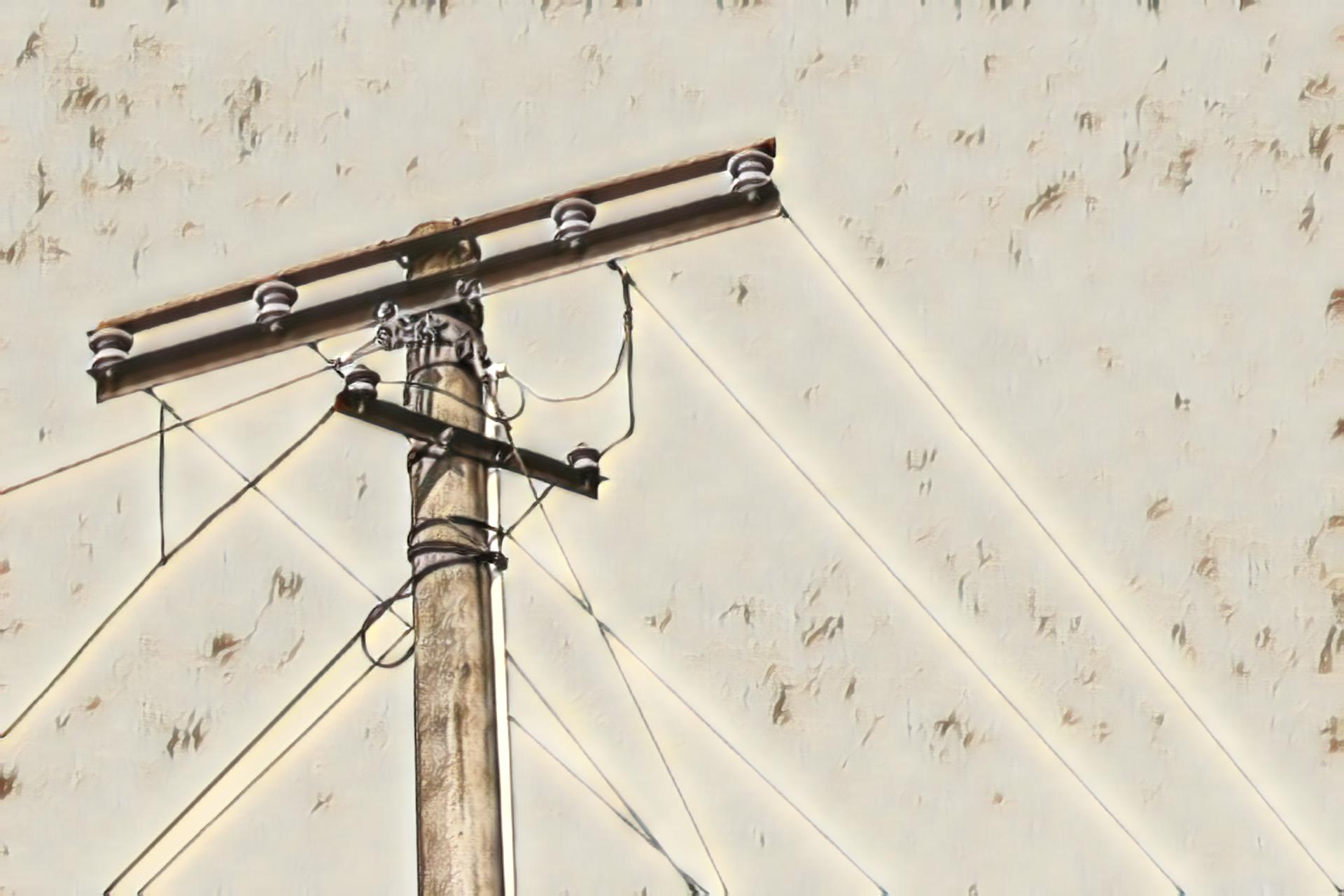} & \includegraphics[width=0.155\textwidth]{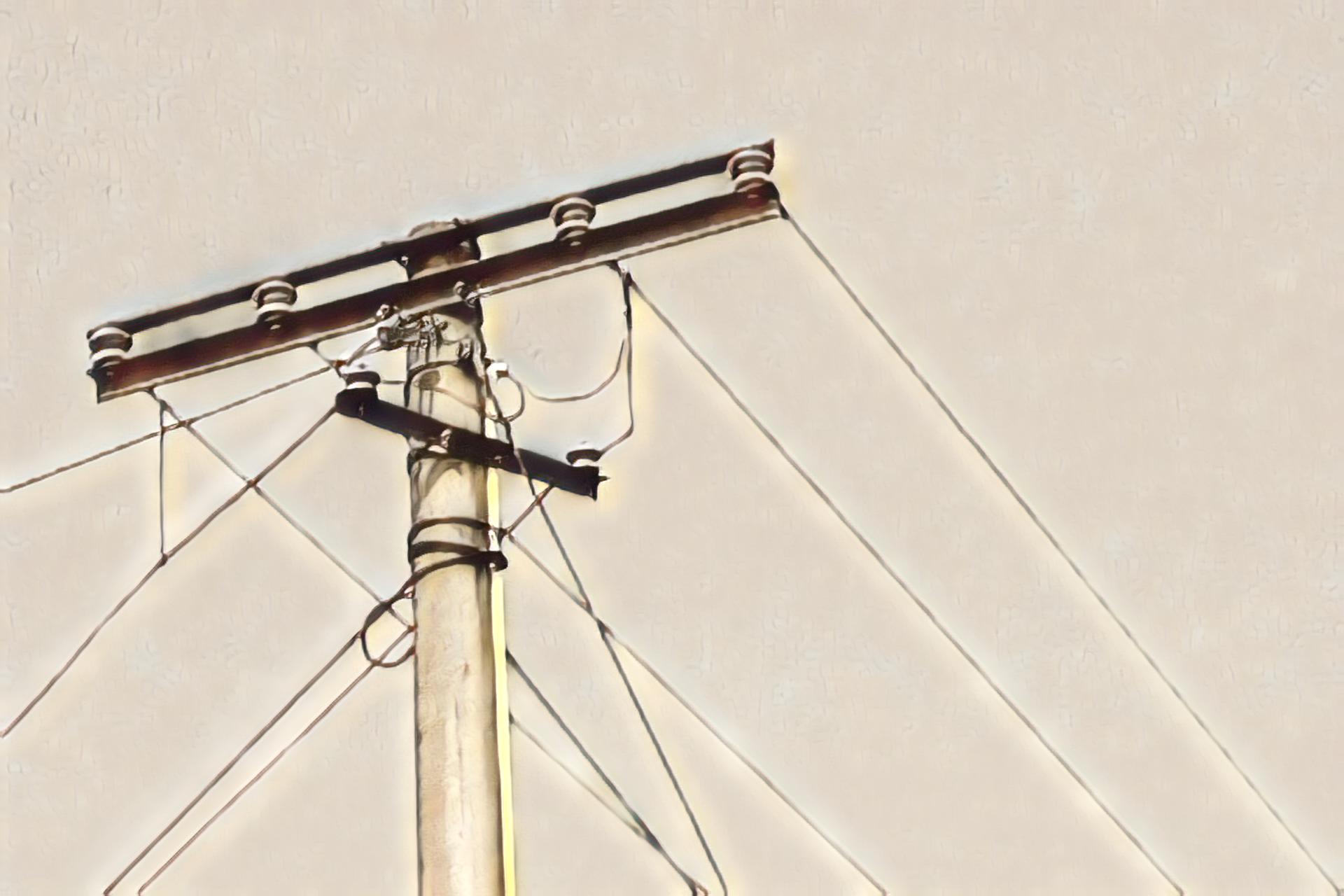}\\
\includegraphics[width=0.155\textwidth]{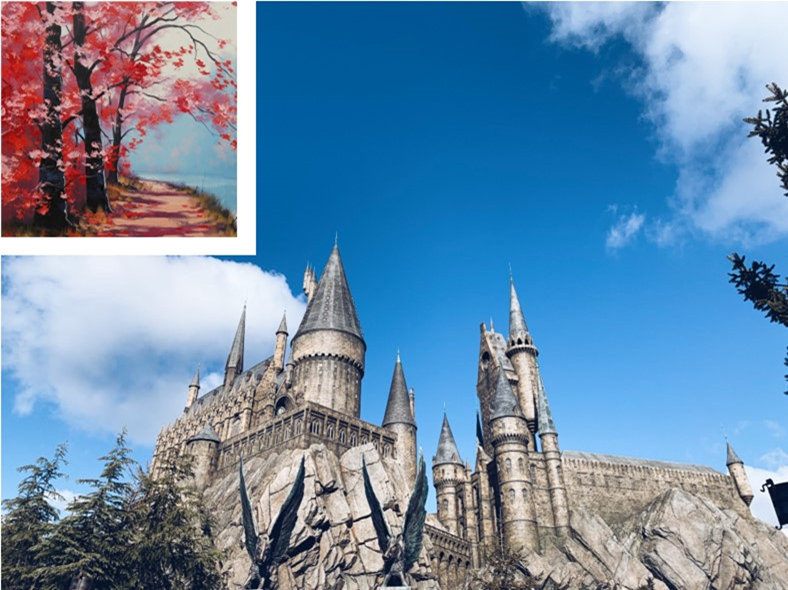}& \includegraphics[width=0.155\textwidth]{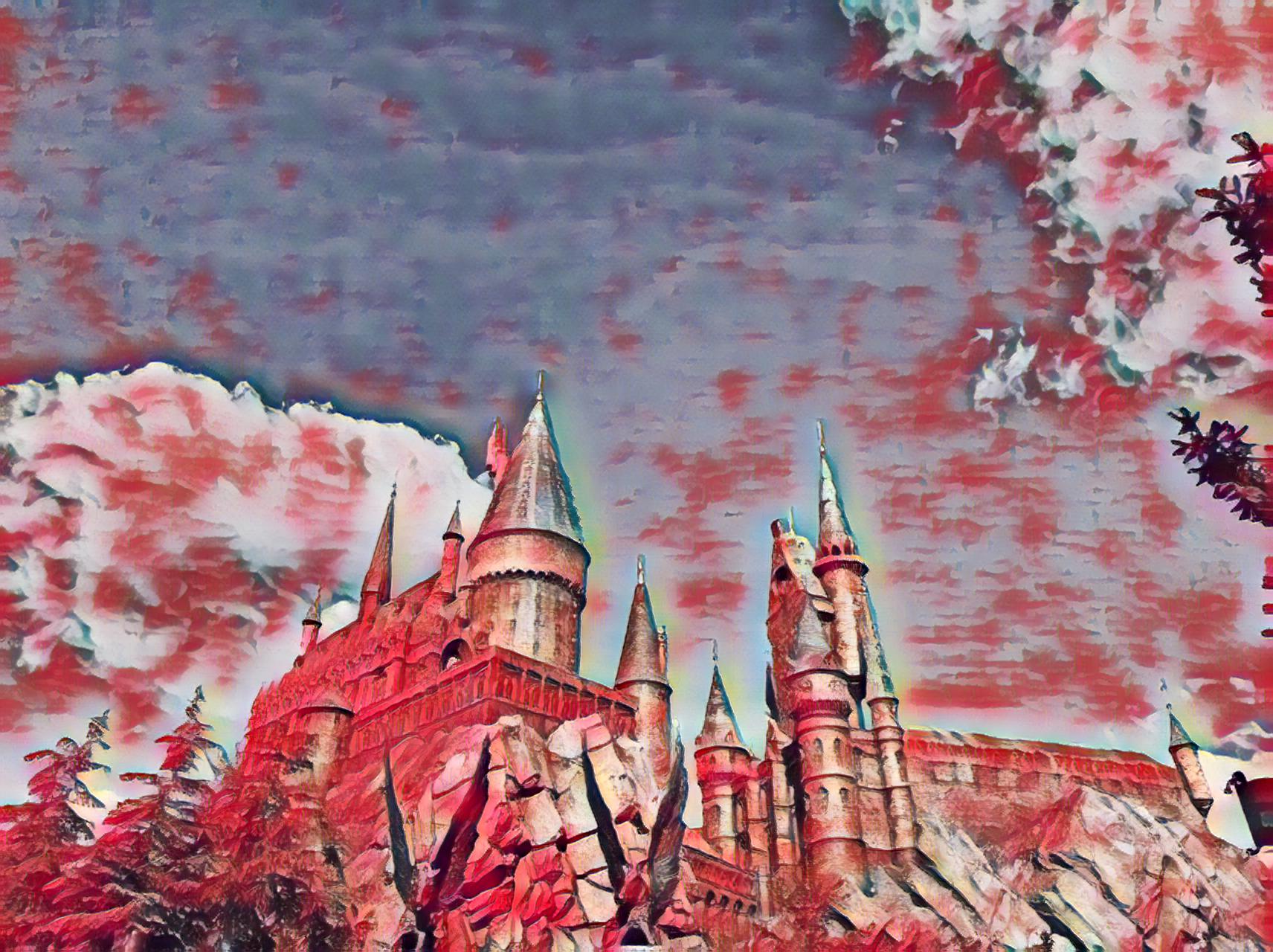} & \includegraphics[width=0.155\textwidth]{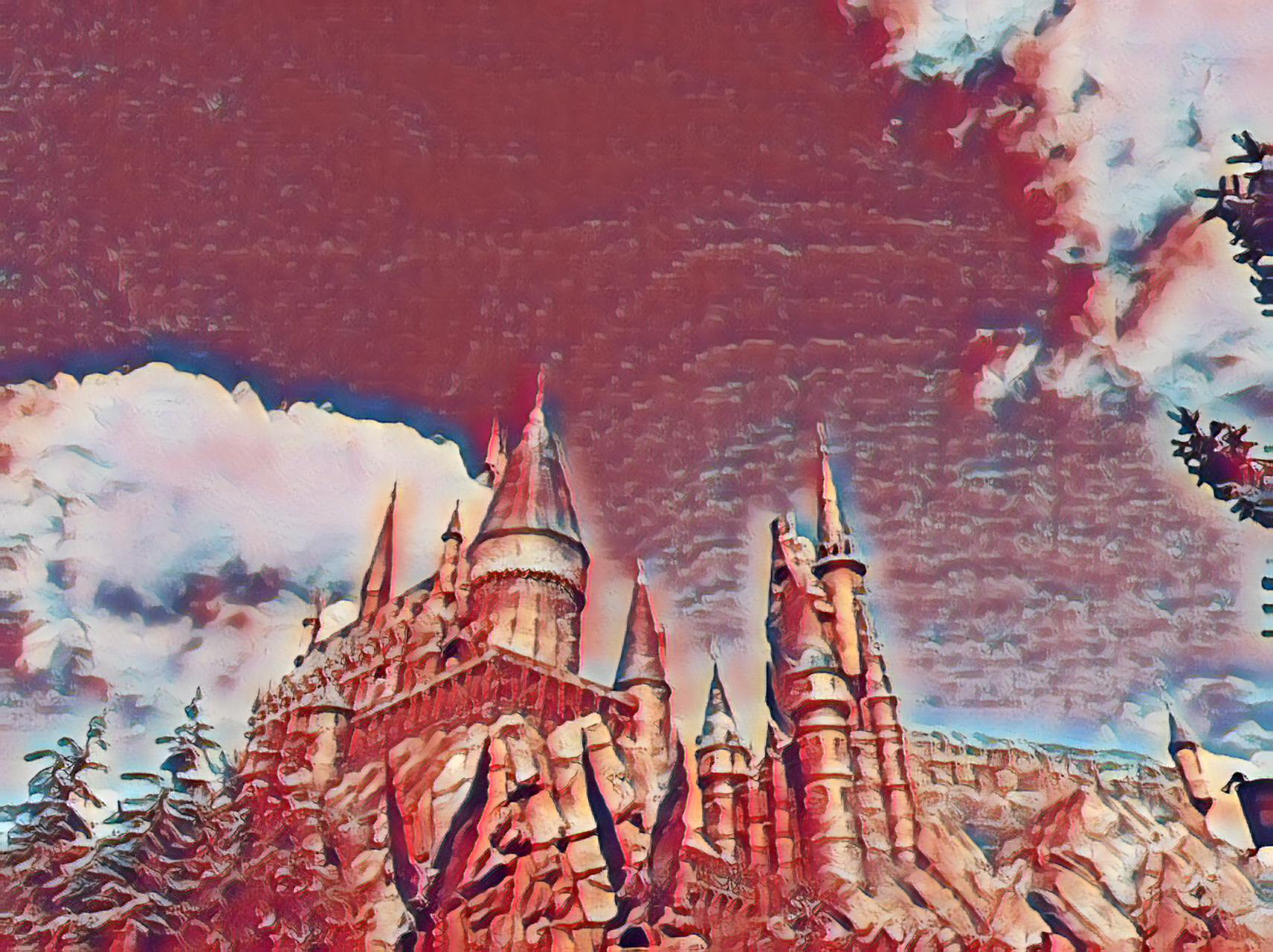} \\

\small{Content \& Style}&\small{Standard DIN} & \small{Deformable DIN}
\end{tabular}

}
\vspace{-0.1cm}
%\vspace{-0.01cm}
\caption{Comparative results of the standard DIN and one of its variants, deformable DIN.}
\label{fig:deformable} %% label for entire figure
\end{figure}

\begin{table}[!t]
\caption{Computation complexities (GFLOPs) of different network architectures with input size of $512 \times 512$.}
\begin{center}
\setlength\tabcolsep{12 pt}
{\renewcommand{\arraystretch}{1.3}
\begin{tabular}{lccc}
  \noalign{\hrule height 0.8pt}

  \multirow{2}{*}{\textbf{Methods}} &\multicolumn{2}{c}{\textbf{Complexity}}\\ \cline{2-3} & \textbf{Encoder} & \textbf{Decoder}\\
  %\textbf{Methods} & \textbf{Encoder} & \textbf{Decoder} \\
  \noalign{\hrule height 0.5pt}
  Johnson \etal & 14.16 & 19.59 \\
  Li \etal & 203.42 & 203.10 \\
  Huang \etal & 63.44 & 63.36 \\
  Sheng \etal & 63.44 & 63.36 \\
  Li and Liu \etal & 63.44 & 63.36 \\
  Ours (MobileNet) & \textbf{3.62} & \textbf{3.72} \\
  \noalign{\hrule height 0.8pt}
\end{tabular}}
\end{center}
\label{tab:flops}
\end{table}

%-------------------------------------------------------------------------------|---------------------------------------
\subsection{Qualitative Evaluation}
%-------------------------------------------------------------------------------|---------------------------------------

Fig.~\ref{fig:quality} demonstrates the qualitative results of the proposed DIN and other state-of-the-art ASPM algorithms \cite{li2017universal,huang2017arbitrary,sheng2018neural}, which also use one single model for arbitrary style transfers.
The comparison results are produced by using the official implementations with the default settings provided by the authors.
All the testing content and style images are not used in training.
In the last column of Fig.~\ref{fig:quality}, we also produce the corresponding stylization results by using the popular PSPM algorithm of \cite{Johnson2016perceptual} as a gold standard, which trains separate style-specific models for different styles.
In particular, to validate the effectiveness of the proposed DIN layer, we produce both the results of using the same VGG encoder as other ASPM algorithms (\emph{Ours (VGG)}), and also the results of the proposed hierarchical MobileNet-based network (\emph{Ours (MobileNet)}), which is about twenty times smaller than the VGG encoder.

The algorithm of Li \textit{et al}., as shown in Fig.~\ref{fig:quality}, is not good at generating sharp details and fine strokes, due to its learning-free manner.
Their results generally have distorted style patterns, \eg, the background clutters in the \nth{5} column of Fig.~\ref{fig:quality}.
By contrast, the algorithm of Huang \etal generates finer strokes; however, their algorithm is not effective at handling challenging style patterns, which is especially obvious in the \nth{1} row, \nth{6} column of Fig.~\ref{fig:quality}, where very few style patterns are transferred.
Also, their results have some similar repeated texture patterns among different styles, which might be caused by the suboptimal manually defined way for calculating parameters in AdaIN layer.
The results of Sheng \etal also suffer from the issue of distorted patterns and lacking details.
By contrast, with the same VGG encoder, our proposed DIN demonstrates superior performance in transferring challenging style patterns and meanwhile producing finer details (Fig.~\ref{fig:quality}, \nth{3} column).
Even if the encoder is reduced by a factor of twenty in complexity, the proposed DIN still achieves comparable quality (Fig.~\ref{fig:quality}, \nth{4} column) to the gold-standard PSPM algorithm of Johnson \etal (Fig.~\ref{fig:quality}, \nth{8} column).

\begin{figure}[!t]
\setlength\tabcolsep{0 pt}
{\renewcommand{\arraystretch}{0.7}
%\begin{tabular}{>{\centering}n{\p} >{\centering}n{\p} >{\centering\arraybackslash}n{\p}}
\begin{tabular}{>{\centering}m{2.13cm} >{\centering}m{2.13cm} >{\centering}m{2.13cm} >{\centering\arraybackslash}m{2.13cm}}
\centering
%\begin{tabular}{cccc}
%\centering

\includegraphics[width=0.115\textwidth]{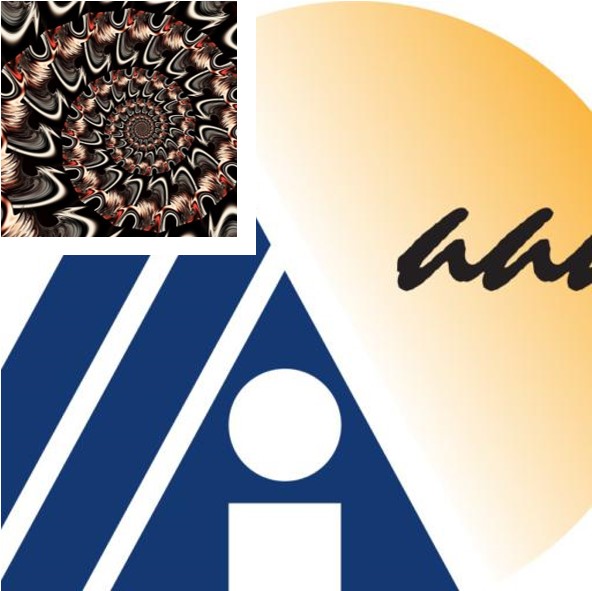}&\includegraphics[width=0.115\textwidth]{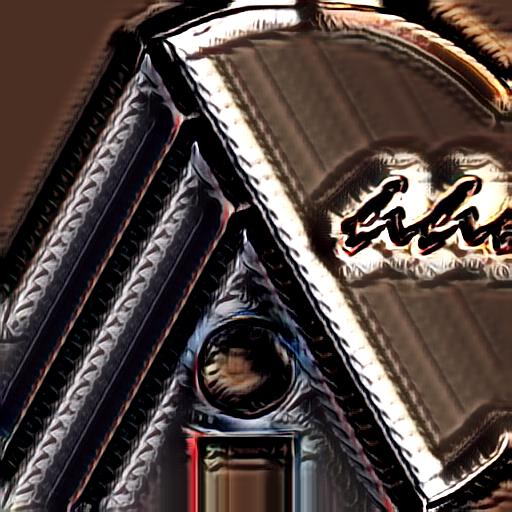}& \includegraphics[width=0.115\textwidth]{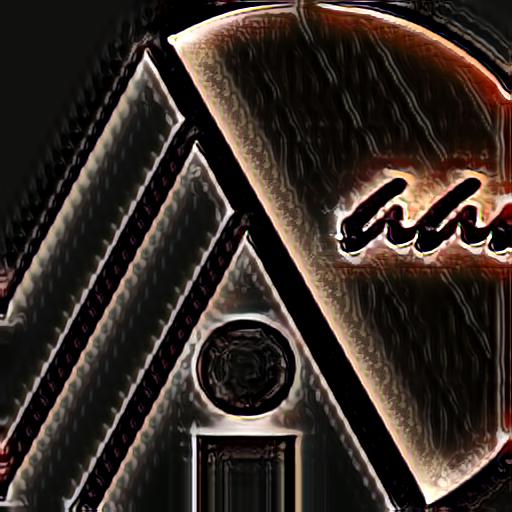} & \includegraphics[width=0.115\textwidth]{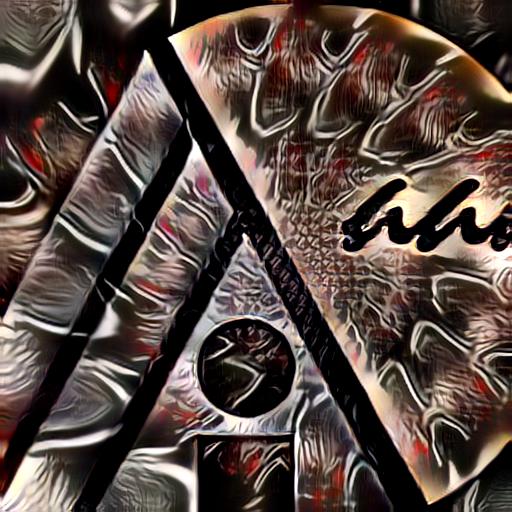} \\

\includegraphics[width=0.115\textwidth]{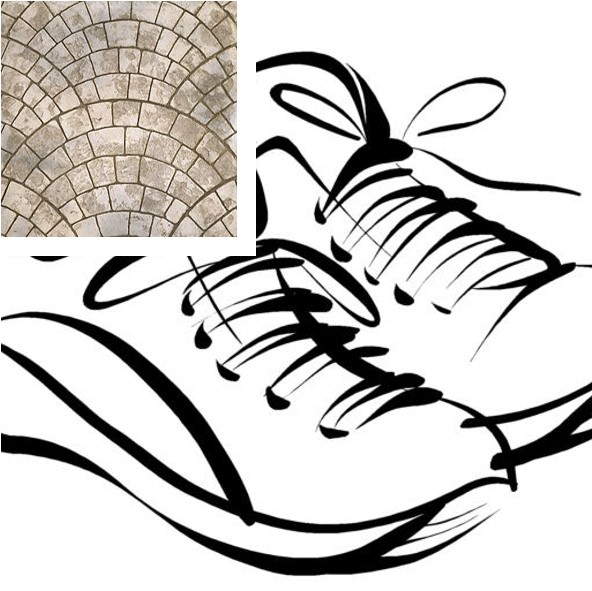}&\includegraphics[width=0.115\textwidth]{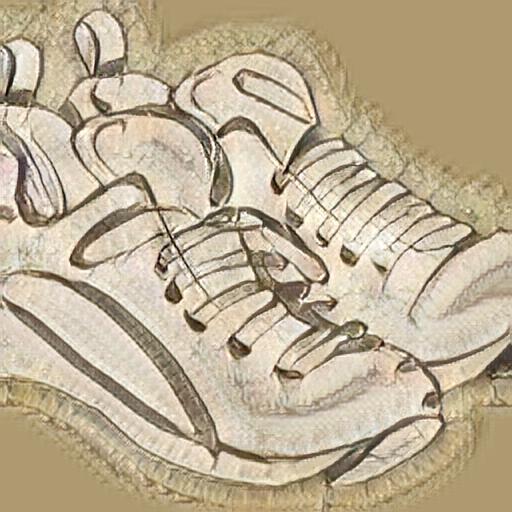}& \includegraphics[width=0.115\textwidth]{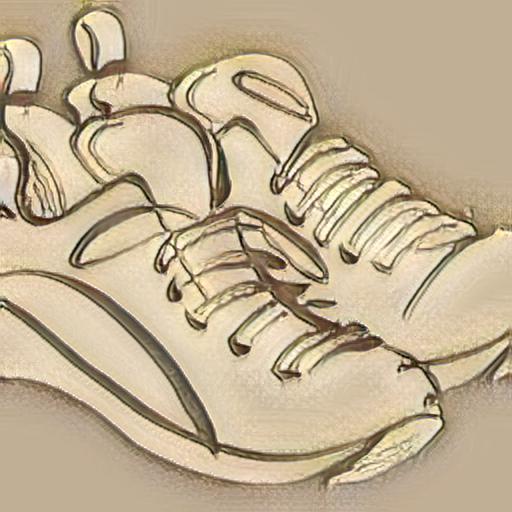} & \includegraphics[width=0.115\textwidth]{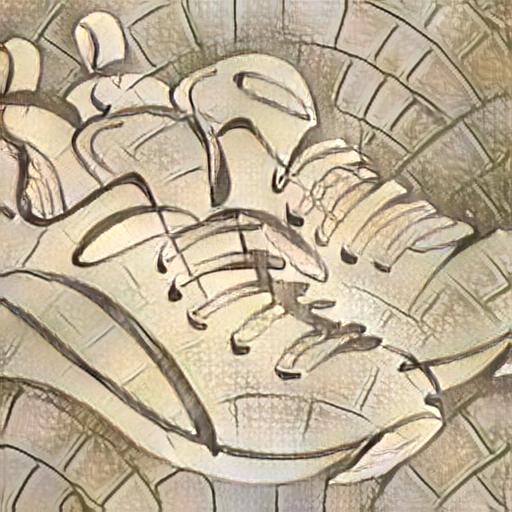} \\

%\smallskip
\scriptsize{\vspace{-0.24cm}Content \& Style}&\scriptsize{\vspace{-0.24cm}AdaIN} & \scriptsize{\vspace{-0.24cm}Standard DIN} & \multirow{1}{2.7cm}{\scriptsize{Spatially-adaptive DIN}}

%\scriptsize{(a)}&\scriptsize{(b)} & \scriptsize{(c)} & \scriptsize{(d)}
\end{tabular}

}
\vspace{-0.4cm}
%\vspace{-0.01cm}
\caption{Comparative results of non-natural images produced by our standard DIN and the proposed spatially-adaptive DIN.}
\label{fig:spade} %% label for entire figure
\end{figure}

Also, we show in Fig.~\ref{fig:deformable} and Fig.~\ref{fig:spade} the results of the two aforementioned variants of the proposed DIN, respectively.
Compared with standard DIN that uses standard convolutions, the proposed deformable DIN achieves automatic spatial-stroke control according to the visual attention, thus avoiding random stroke placement for fore- and background objects (\eg, black and red strokes in the \nth{2} column of Fig.~\ref{fig:deformable}).
The other variant, spatially-adaptive DIN, can generate proper strokes for uniform pixel areas in non-natural images, as shown in the last column of Fig.~\ref{fig:spade}.

\begin{figure}[!t]
  \centering
  \includegraphics[width=0.45\textwidth]{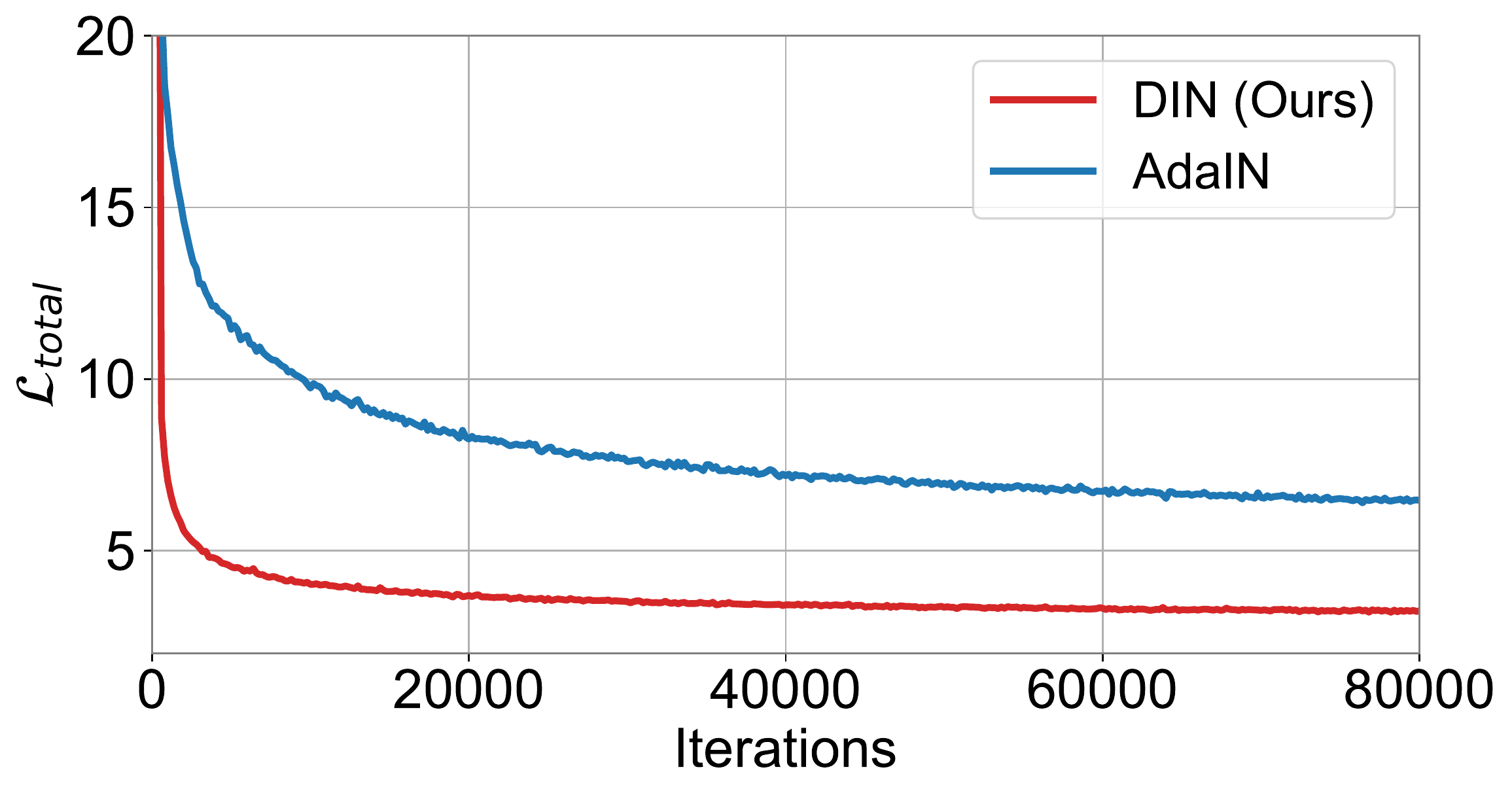}\\
  \caption{Comparing the {training} curves of the proposed DIN and AdaIN with the lightweight network architecture.}\label{fig:trainingcurve}
\end{figure}

%-------------------------------------------------------------------------------|---------------------------------------
\subsection{Quantitative Evaluation}
%-------------------------------------------------------------------------------|---------------------------------------

Tab.~\ref{tab:flops} shows the comparison results among different architectures in terms of computation complexity, \ie, the number of floating-point operations.
Our proposed MobileNet-based network architecture outperforms other arbitrary stylization networks by a large margin in efficiency, even having much less computational cost than the PSPM network of \cite{Johnson2016perceptual}.
In particular, the additional weight and bias generators in our architecture bring little extra cost, \ie, only 19.97MFLOPs given $512 \times 512$ input.

Also, to demonstrate the proposed DIN can lead to a better optimization,  we show  in Fig.~\ref{fig:trainingcurve} the quantitative comparison results in terms of training curves.
%
%It is noticeable that under the same settings, our proposed DIN achieves faster and better convergence.
Our proposed DIN, as can be noticed, achieves faster and better convergence under the same settings.
%than AdaIN .

%their idea of multi-level stylization needs five encoders and decoders

%===============================================================================|=======================================
\subsection{Ablation Study}\label{subsection:AblationStudy}
%===============================================================================|=======================================

\begin{figure}[!t]
\setlength\tabcolsep{1 pt}
{\renewcommand{\arraystretch}{0.8}
%\begin{tabular}{>{\centering}n{\p} >{\centering}n{\p} >{\centering\arraybackslash}n{\p}}
%\begin{tabular}{>{\centering}m{1.98cm} >{\centering}m{1.98cm} >{\centering}m{1.98cm} ?{0.2mm} >{\centering}m{1.98cm} >{\centering}m{1.98cm} >{\centering\arraybackslash}m{1.98cm}}
\centering
%\begin{tabular}{cccc}
%\ \ \qquad&\includegraphics[width=0.14\textwidth]{figs/mobilenet/content1.jpg}& \includegraphics[width=0.14\textwidth]{figs/mobilenet/without1.jpg} &\includegraphics[width=0.14\textwidth]{figs/mobilenet/with1.jpg}\\
%
%
%
%%\ \ \qquad&\includegraphics[width=0.14\textwidth]{figs/mobilenet/content2.jpg}& \includegraphics[width=0.14\textwidth]{figs/mobilenet/without2.jpg} &\includegraphics[width=0.14\textwidth]{figs/mobilenet/with2.jpg}\\
%
%\smallskip
%\ \ \qquad&\small{(a) Content \& Style}& \small{(b) Depthwise Conv} & \small{(c) Standard Conv}
%%\scriptsize{(a)}&\scriptsize{(b)} & \scriptsize{(c)} & \scriptsize{(d)}
%\end{tabular}

\begin{tabular}{cccc}
\centering

%\includegraphics[width=0.119\textwidth]{figs/mobilenet/content1.jpg}& \includegraphics[width=0.119\textwidth]{figs/mobilenet/without1.jpg} & \includegraphics[width=0.119\textwidth]{figs/mobilenet/with1.jpg} &\includegraphics[width=0.119\textwidth]{figs/mobilenet/adain.jpg}\\
%
%\smallskip
%\scriptsize{(a) Content \& Style}&\scriptsize{(b) Depthwise Conv} & \scriptsize{(c) Standard Conv} & \scriptsize{(d) AdaIN}\\
%
%\includegraphics[width=0.119\textwidth]{figs/unet/content.jpg}& \includegraphics[width=0.119\textwidth]{figs/unet/with.jpg} & \includegraphics[width=0.119\textwidth]{figs/unet/without.jpg} &\includegraphics[width=0.119\textwidth]{figs/unet/adain.jpg}\\
%
%\smallskip
%\scriptsize{(e) Content \& Style}&\scriptsize{(f) Hierarchical DIN} & \scriptsize{(g) Single-level DIN} & \scriptsize{(h) AdaIN}

\includegraphics[width=0.115\textwidth]{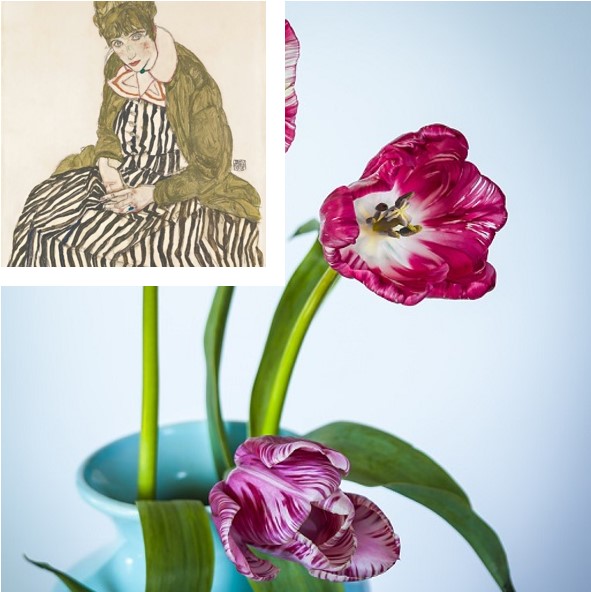}&\includegraphics[width=0.115\textwidth]{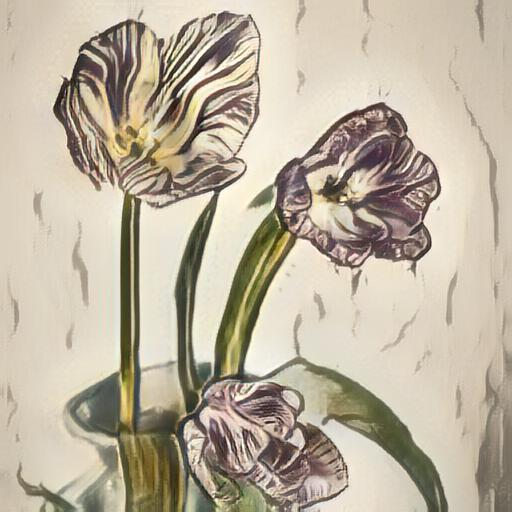}& \includegraphics[width=0.115\textwidth]{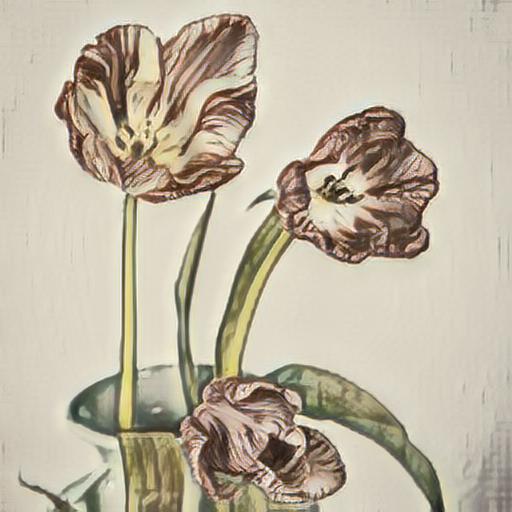} & \includegraphics[width=0.115\textwidth]{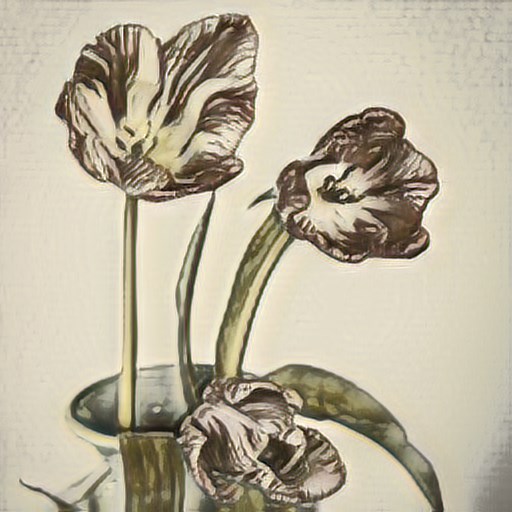} \\

\smallskip
\scriptsize{(a) Content \& Style}&\scriptsize{(b) AdaIN} & \scriptsize{(c) Standard Conv} & \scriptsize{(d) Depthwise Conv}\\

\includegraphics[width=0.115\textwidth]{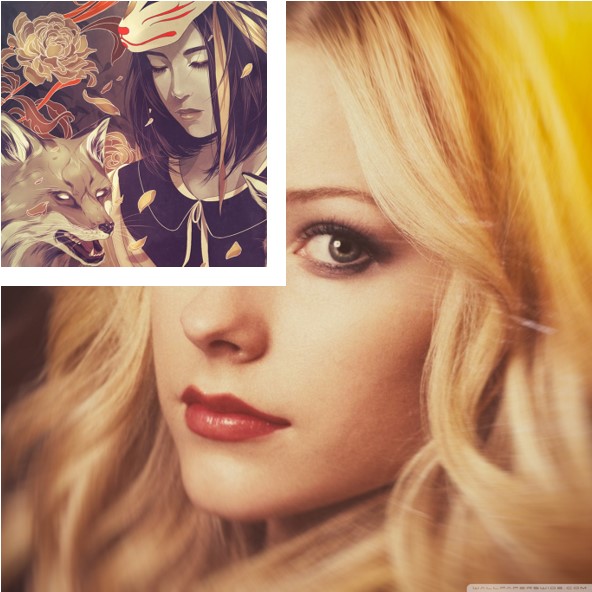}&\includegraphics[width=0.115\textwidth]{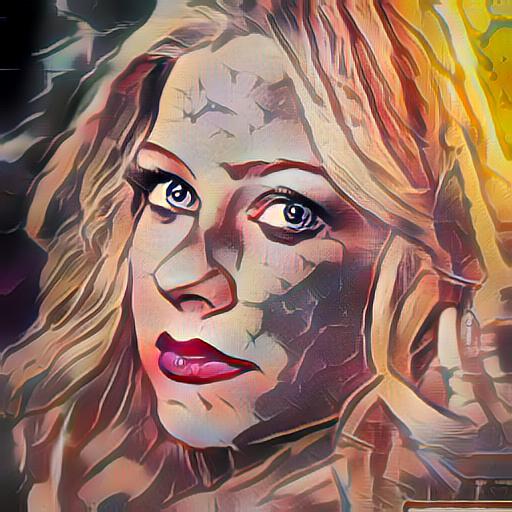}& \includegraphics[width=0.115\textwidth]{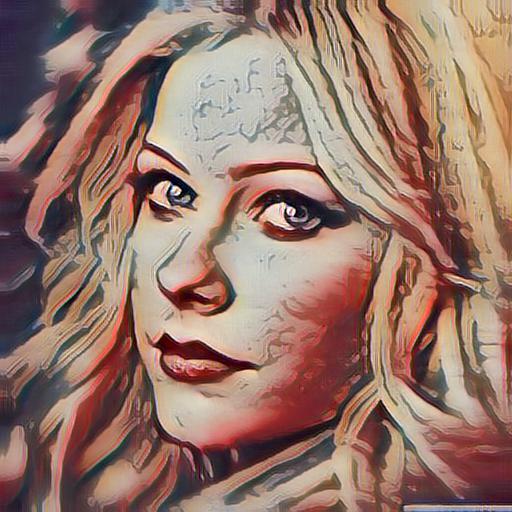} & \includegraphics[width=0.115\textwidth]{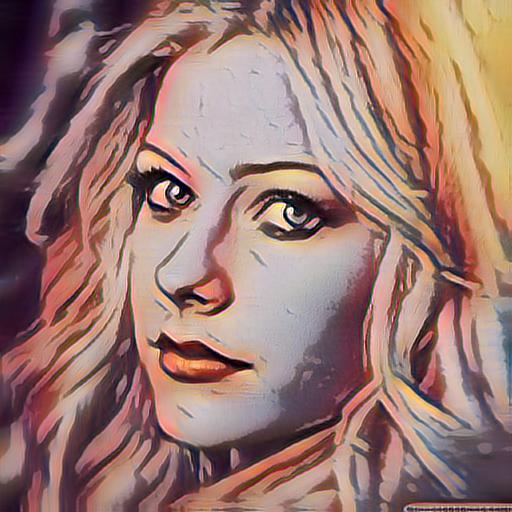} \\

\smallskip
\scriptsize{(e) Content \& Style}&\scriptsize{(f) AdaIN} & \scriptsize{(g) Single-level DIN} & \scriptsize{(h) Hierarchical DIN}

%\scriptsize{(a)}&\scriptsize{(b)} & \scriptsize{(c)} & \scriptsize{(d)}
\end{tabular}

}
\vspace{-0.2cm}
%\vspace{-0.01cm}
\caption{Results obtained using (c) standard convolutions, (d) depthwise convolutions, (g) single-level DIN, and (h) hierarchical DIN {in the network architecture of Fig.}~\ref{fig:arch}, respectively.}
\label{fig:ablation} %% label for entire figure
\vspace{-0.2cm}
\end{figure}

To validate the proposed lightweight architecture in Fig.~\ref{fig:arch}, we perform extensive ablation studies and demonstrate the corresponding results in Fig.~\ref{fig:ablation}.
The first row of Fig.~\ref{fig:ablation} shows the stylization results of using standard convolutions and depthwise separable convolutions in our network architecture.
Despite the lower computational complexity, the visual quality using depthwise separable convolutions is still comparable to that using standard convolutions.
In the second row of Fig.~\ref{fig:ablation}, we compare the results of using one single DIN layer and multiple hierarchical DIN layers depicted in Fig.~\ref{fig:arch}.
Our hierarchical design preserves finer structures of the input image, as can be observed in the human face of Fig.~\ref{fig:ablation}(h).
Additional ablation studies including varying kernel sizes can be found in the supplementary material.

%with mobilenet and without mobilenet
%
%\noindent\textbf{Small kernel size \vs large kernel size in DIN layer.}
%kernel size is 3 vs kernel size is 1
%
%\noindent\textbf{Group-wise convolution \vs non-group-wise convolution in DIN layer.}
%group-wise for DIN layer
%based on unet or based on vgg?
%
%
%\noindent\textbf{With and Without Unet}

%===============================================================================|=======================================
\section{Discussions and Conclusions}\label{section:Conclusion}
%===============================================================================|=======================================

%\jyc{do not need to align feature mean and variance}
%use affine transformation and manually define the way to compute the parameters, we model the stylization process as a more generalized dynamic convolutional transformation, of which the parameters are adaptively updated in a learnable manner.
%
%The proposed DIN enables a sophisticated style encoder to adequately encode challenging styles, yet comes with a lightweight content encoder.

In this paper, we introduce a novel dynamic instance
normalization layer (DIN) for flexible and more efficient arbitrary style transfers.
The proposed DIN allows for the use of an elaborate style encoder
to encode rich style patterns and a lightweight content encoder for improved efficiency, thereby resolving the dilemma of redundant and shared encoders in previous methods.
%
%As a result, the proposed DIN
%yields encouraging results in transferring challenging style patterns, and meanwhile achieves a speedup factor of more than twenty as compared to the state of the art.
%
Experimental results demonstrate that the proposed approach achieves favorable performance against the state of the art, especially in transferring challenging style patterns
while preserving a very light computational cost.
%, as compared with previous work.
%
In addition, by incorporating state-of-the-art convolutional operations,
the proposed DIN is able to create novel effects in arbitrary style transfers, such as
automatic spatial-stroke control.
In our future work, we would like to explore the use of DIN
in other computer vision tasks beyond style transfer~\cite{huang2018multimodal,wang2011subspaces,dundar2018domain,wang2014tracking,liu2019few},
as the proposed DIN can  readily replace
other normalization layers like CIN and AdaIN.
In particular, we would like to introduce DIN to the context of domain adaption~\cite{dundar2018domain},
since style transfer is intrinsically a domain-adaption task~\cite{li2017demystifying,li2019learning}.

\subsubsection{Acknowledgement.}
This work is supported by  National Key Research and Development Program (2016YFB1200203), National Natural Science Foundation of China (61976186),  Key Research and Development Program of Zhejiang Province (2018C01004), and the Major Scientifc Research Project of Zhejiang Lab (No. 2019KD0AC01) .
Xinchao Wang is supported by the startup funding of Stevens Institute of Technology.

\bibliographystyle{aaai} \bibliography{MYRE}
\end{document}